# Robust 6DoF Pose Tracking Considering Contour and Interior Correspondence Uncertainty for AR Assembly Guidance

Jixiang Chen, Jing Chen, *Member, IEEE*, Kai Liu, Haochen Chang, Shanfeng Fu, and Jian Yang

*Abstract*—**Augmented reality assembly guidance is essential for intelligent manufacturing and medical applications, requiring continuous measurement of the 6DoF poses of manipulated objects. Although current tracking methods have made significant advancements in accuracy and efficiency, they still face challenges in robustness when dealing with cluttered backgrounds, rotationally symmetric objects, and noisy sequences. In this paper, we first propose a robust contour-based pose tracking method that addresses error-prone contour correspondences and improves noise tolerance. It utilizes a fan-shaped search strategy to refine correspondences and models local contour shape and noise uncertainty as mixed probability distribution, resulting in a highly robust contour energy function. Secondly, we introduce a CPU-only strategy to better track rotationally symmetric objects and assist the contour-based method in overcoming local minima by exploring sparse interior correspondences. This is achieved by pre-sampling interior points from sparse viewpoint templates offline and using the DIS optical flow algorithm to compute their correspondences during tracking. Finally, we formulate a unified energy function to fuse contour and interior information, which is solvable using a re-weighted least squares algorithm. Experiments on public datasets and real scenarios demonstrate that our method significantly outperforms state-of-the-art monocular tracking methods and can achieve more than 100 FPS using only a CPU.**

*Index Terms*—**6DoF pose tracking, vision-based measurement, probability model, CPU-only strategy, AR assembly guidance.**

## I. Introduction

Augmented reality (AR) assembly guidance is a promising research field that is widely applied in everyday life [1], intelligent manufacturing [2], and medical scenarios [3]. By utilizing optical see-through head-mounted displays (OST HMDs), 3D virtual information can be seamlessly integrated with real-world objects, providing visual feedback that effectively enhances work efficiency and assists operators in making subsequent decisions [4], [5]. To achieve better fusion of virtual and real elements, thereby offering users a more immersive experience, real-time, continuous, and accurate measurement of the 6DoF pose of 3D objects is crucial.

Numerous 6DoF object pose measurement methods have been proposed, including various sensor measurement techniques such as optical trackers [5], electromagnetic trackers [6], ultrasound trackers [7], and cameras. These methods can be categorized as: feature-point-based [8]-[12], direct [13]-[16], edge-based [17]-[22], region-based [23]-[30], multi-feature-based [31]-[41], and deep learning-based [42]-[45] object tracking methods. Among these, monocular 6DoF object tracking methods are low-cost and easy to deploy, making them particularly suitable for environments with constrained computational resources. Therefore, in this paper, we mainly focus on monocular 6DoF pose tracking methods.

In AR-based industrial assembly environments, the manipulated objects are typically texture-less and rotationally symmetric, and the working conditions are complex, involving cluttered backgrounds, occlusion, and image noise. These factors make robust tracking particularly challenging. Edge-based and region-based methods are suitable for tracking texture-less objects, as both aim to find an optimal pose that best aligns projected contours with the object contours. Therefore, we categorize them as contour-based methods. Recently, these methods have seen significant improvements in both accuracy and efficiency. To tackle the issue of partial occlusions of objects and the ambiguous colors present in backgrounds, [27] proposed a contour point search strategy considering distance and color constraint along with a pixel-wise weighted energy function. To better track less-distinct objects, [35] proposed a contour-part model which detects correspondences by gradient-orientation-based template matching. To improve efficiency, [28] proposed a correspondence line model and new smooth step functions that generate line-wise posterior probabilities with Gaussian property. In addition, sparse viewpoint model was introduced to improve efficiency. To further address large displacement issues, [30] proposed a hybrid optimization method that combines local and non-local optimization. While promising, there are still potential improvements when handling complex scenarios in industrial assembly.

Firstly, experimental results on RBOT public dataset [25] demonstrate that, despite achieving state-of-the-art performance, contour-based methods still exhibit unsatisfied tracking capabilities in sequences with additive noises and cluttered backgrounds. Except for the reason that noises and c-

This work was supported in part by Basic Research, under grant JCKY2021602B021 and National Science Foundation Program of China, under grant 62025104. *(Corresponding author: Jing Chen).*

Jixiang Chen, Jing Chen, Kai Liu, Shanfeng Fu, and Jian Yang are with the School of Optics and Photonics, Beijing Institute of Technology, Beijing, 100081, China (e-mails: 3220235081@bit.edu.cn, chen74jing29@bit.edu.cn, 3120245404@bit.edu.cn, 3120230520@bit.edu.cn, jyang@bit.edu.cn).

Haochen Chang is with the School of Systems Science and Engineering, Sun Yat-Sen University, Guangzhou, 510006, China (e-mail: changhch5@mail2.sysu.edu.cn).

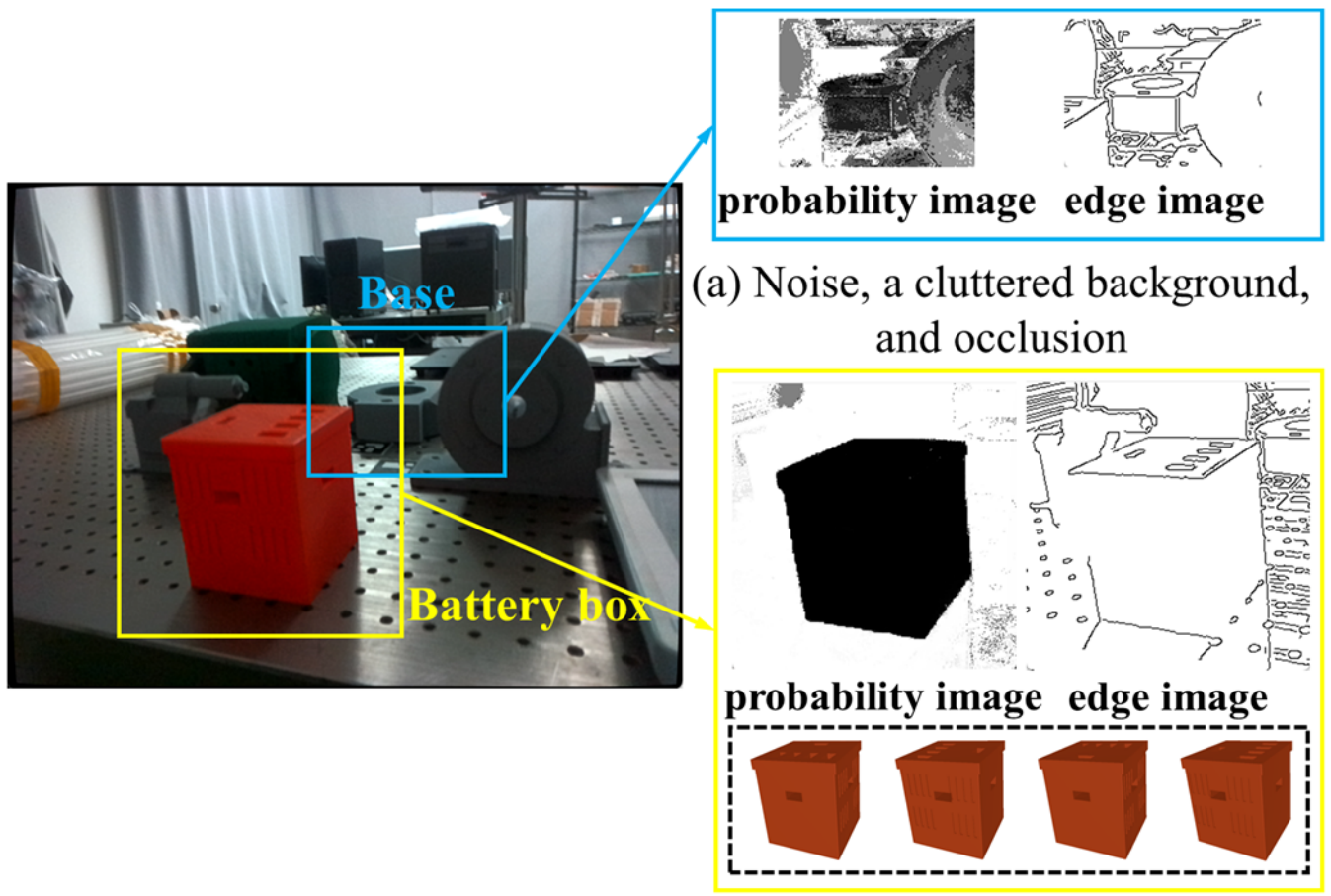

(a) Noise, a cluttered background, and occlusion

(b) Same contours in different poses

**Fig. 1.** Challenging scenarios for object tracking in industrial assembly.

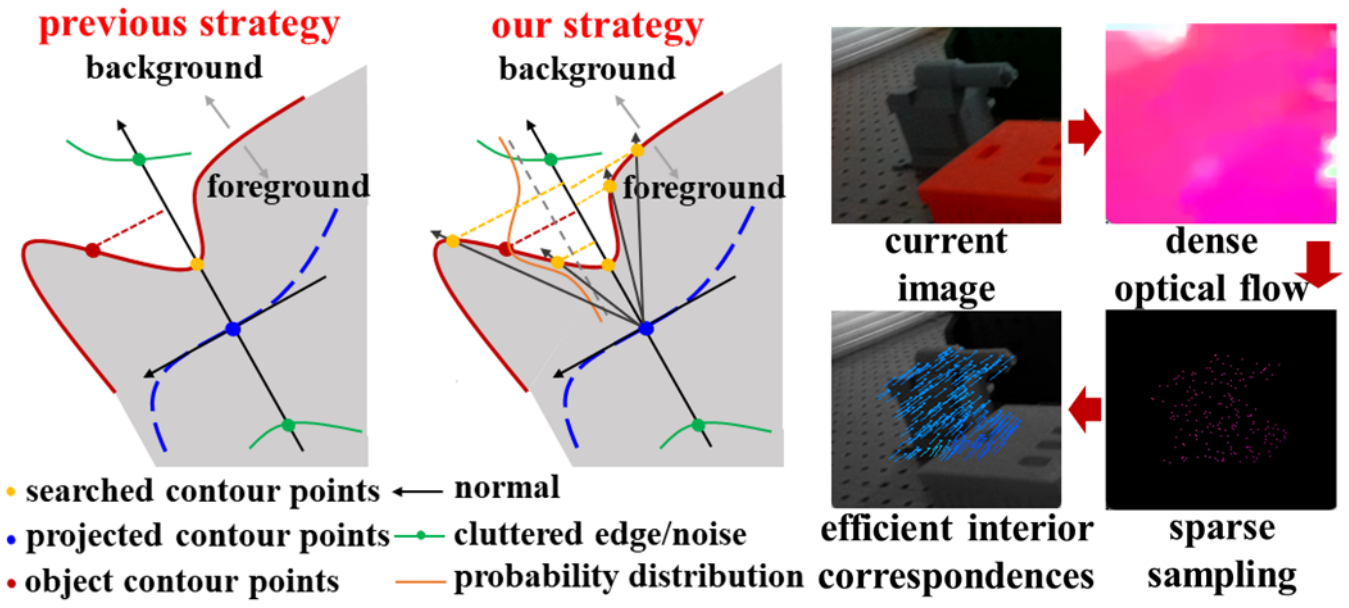

(a) Different contour association strategies (b) Efficient interior association

**Fig. 2.** Proposed data association strategies.

luttered edges can obscure intricate image details, making contour matching more challenging, noise itself may also cause inaccurate color statistics and blur the distinction between foreground and background, as illustrated in Fig. 1(a). Furthermore, contour-based methods normally utilize a line searching strategy to improve efficiency. This strategy assumes that the projected contour and object contour have nearly the same orientation and curvature, thereby enabling the search for correspondences along 1D normal line. However, this assumption is prone to failure when dealing with contours that have large curvature or significant inter-frame pose variation.

Secondly, contour-based methods, which rely on contour alignment, may encounter difficulties in accurately estimating pose due to the ambiguity of contour, as illustrated in Fig. 1(b). Although regularization strategy [28] can help constrain pose variation, it is only effective in cases involving small out-plane rotations. Therefore, this problem is not fundamentally resolved. An alternative approach is to leverage correspondences within the object silhouette between two consecutive frames. To achieve this, feature-point-based methods [8]-[12], direct methods [13]-[16], and optical flow algorithms [41] have been proposed. However, the lack of efficient, robust, and CPU-only implementation strategy restricts their applications. In addition, contour-based methods alone are unsuitable for measuring objects with significant inter-frame pose variation, as the non-convex nature of the energy function makes the estimated pose prone to getting stuck in local minima during optimization. To tackle this issue, combining local and non-local optimization [30] or selecting transitional view [29] has been proposed. Although the tracking success rates are ameliorated to a certain extent, real-time performance and additional device dependency still limit their applications.

Based on above observations, we propose a novel correspondence searching strategy, named as point-to-distribution searching, to achieve robust and efficient monocular contour-based pose tracking. Our approach distinguishes itself from previous works by utilizing a fan-shaped strategy to search potential object contour points and then project these points onto the normal of precomputed projected contour point to generate a 1D probability distribution. The mean of this distribution is considered as the most likely position of the correspondence, while the variance quantifies the uncertainty in that position and implicitly reflects the effects of noise and the shape complexity of the local object contour. To model this distribution, we introduce a mixed probability distribution composed of a Gaussian distribution and a uniform distribution. This mixed distribution enhances the robustness of contour energy, correspondences with high uncertainty can be effectively suppressed during optimization. The differences between previous methods and ours are shown in Fig. 2(a).

To make our tracking method robust for rotationally symmetric objects, we also propose a CPU-only strategy that leverages sparse interior correspondences of the object. Specifically, we randomly pre-sample interior points in viewpoint templates offline and purely use CPU to compute their corresponding points by DIS algorithm [46] online, as depicted in Fig. 2(b). Unlike traditional methods that rely on feature points [12], our approach allows us to pre-define arbitrary number of interior points at any position and compute their correspondences. Based on these sparse correspondences, we particularly establish a robust interior energy incorporating extended patch-wise interior correspondence uncertainty to achieve robust estimation.

To further improve the robustness of our algorithm and handle more complex scenarios, we formulate a unified energy function to fuse contour and interior information that can be simply solved by a re-weighted least squares algorithm. A variable weight is introduced to dynamically adjust the ratio between the two energy components. Initially, interior energy is prioritized to aid optimization cross local minima and finally contour energy is dominant to ensure pixel-level accuracy. This coarse-to-fine strategy makes two modalities better complement each other.

We summarize our main contributions as follows:

1. We propose a robust monocular contour-based pose tracking method that employs fan-shaped search strategy to determine point-to-distribution contour correspondences described by a mixed probability distribution, effectively accounting for shape and noise uncertainties of local contours, which significantly improve the performance especially in noise sequences;

2. We introduce an efficient CPU-only strategy to establish

sparse interior correspondences, that is to pre-sample interior points at pre-generated viewpoint model offline and compute their correspondences by DIS optical flow algorithm online;

3. With contour and interior correspondences, we establish a joint energy function that can be easily solved by re-weighted least squares algorithm. Experimental results have demonstrated improved performances on public datasets and in an AR assembly guidance application.

## II. Related Work

Vision-based 6DoF pose tracking is crucial for augmented reality assembly guidance. Numerous 6DoF object pose tracking methods have been proposed, including feature-based [8]-[16], contour-based [17]-[30], multi-feature-based [4]-[7], [31]-[41], and deep learning-based [42]-[45] tracking approaches. In this section, we provide a brief review of these methods.

### *A. Feature-based Methods*

Feature-point-based methods [8]-[12], which use monocular cameras [8]-[11] or calibrated stereo cameras [12], detect key points with distinct features on the surfaces of objects, and build correspondences between 3D model points and 2D image points by matching key points between two consecutive frames. Pose variation thus can be solved by PnP algorithm. However, only well-textured objects can acquire enough feature points, and incorrect correspondences may lead to tracking failure.

Direct methods [13]-[16] align the pixels of object surface on two consecutive frames by minimizing the photometric error, which rely on the strong assumption of photometric constancy and may easily fail in environment with dynamic light. In addition, both feature-point-based methods and direct methods may occur the accumulation of errors during tracking.

### *B. Contour-based Methods*

Both edge-based [17]-[22] and region-based [23]-[30] methods aim to seek an optimal pose so that the projected contours can best align with the object contours in the captured image. As an early edge-based method, RAPID [17] used 1D search lines to sparsely find corresponding points of the projected contour points in image. Following this work, many improved methods have been proposed. [18] and [19] utilized more robust estimators to solve pose. [20] introduced color statistics cue to achieve optimal contour points searching, and this work was further extended by [21] who proposed to use global optimization to constrain the position of adjacent contour points. In cases of cluttered background and partial occlusion, [22] proposed to use edge confidence fusing color and edge distance cues to search contour and build a weighted energy function.

Instead of explicitly searching for contours, the region-based [23]-[30] methods employ color statistics to segment object and obtain the most likely contours. PWP3D [23] established pixel-wise posterior probabilities and jointly executed image segmentation and pose optimization. [24] proposed to use local color histograms instead of global ones to better track heterogeneous objects. To further improve the robustness of color statistics and optimization strategy, [25] introduced temporally consistent local color histograms attached to model vertexes and a Gauss-Newton-like pose optimization strategy. [26] defined overlapping fan-shaped local regions and proposed to detect occlusion by edge distance and color cues. [27] proposed an optimal contour points searching strategy and a pixel-wise weighted energy function to tackle cases of partial occlusions and ambiguous colors. Recently, [28] developed SRT3D, a sparse region-based 6DoF tracking method which significantly improves accuracy and efficiency. To address large pose shift problem, [29] proposed to apply transitional view to reduce interference of local minima in optimization. Issues such as incorrect correspondences, significant pose displacement, and the presence of symmetrical objects can easily lead to pose measurement failures.

### *C. Multi-feature-based Methods*

Fusing region and edge information is demonstrated better performances in complex scenes. [35] proposed contour-part model that leveraged gradient of edges and region to improve robustness when tracking less-distinct objects. [36] proposed adaptive weighting technique to better balance region and edge information. [37] employed a color statistic model and a geometric edge model to reduce the effect of visual ambiguity during establishing feature correspondences.

Besides the edge and contour, interior region of an object can provide raw pixel data, which also serves as an important visual cue for object pose estimation, and have been combined with region-based methods [38], [39], [41]. For example, [38] and [39] proposed to directly optimize gradient-based descriptor fields [13] to enhance performance based on region-based method [25]. Optical flow algorithms, which can compute 2D correspondences, have also been fused in edge-based method [40]. Later, [41] proposed to fuse region, SIFT feature and optical flow for rigid body and articulated object tracking. However, current methods that utilize interior information of objects require the use of GPUs for rendering during tracking, resulting in significant time consumption and hardware resource usage. Therefore, how to further improve robustness, efficiency and reduce device dependency of the utilization of interior correspondences are still open problems and influence application scenarios.

The tracker that utilizes multi-sensor information fusion has demonstrated excellent performance. When both RGB and depth cameras are available, methods that combine region and depth information often yield outstanding results and have been extensively researched [31]-[34]. For example, [34] introduced a highly efficient and robust probabilistic method ICG, which integrates depth data into a region-based tracking algorithm, enabling it to manage partial occlusions effectively. Other approaches, such as [5], combined Time-of-Flight (ToF) cameras, infrared (IR) cameras, and inertial measurement units (IMUs), utilizing the Unscented Kalman Filter (UKF) to enhance pose tracking. Additionally, [7] proposed an optoacoustic and inertial pose tracking method for moving ob-jects using particle filtering. While these methods show promise, their reliance on specific devices can limit their broader applicability, particularly in industrial settings.

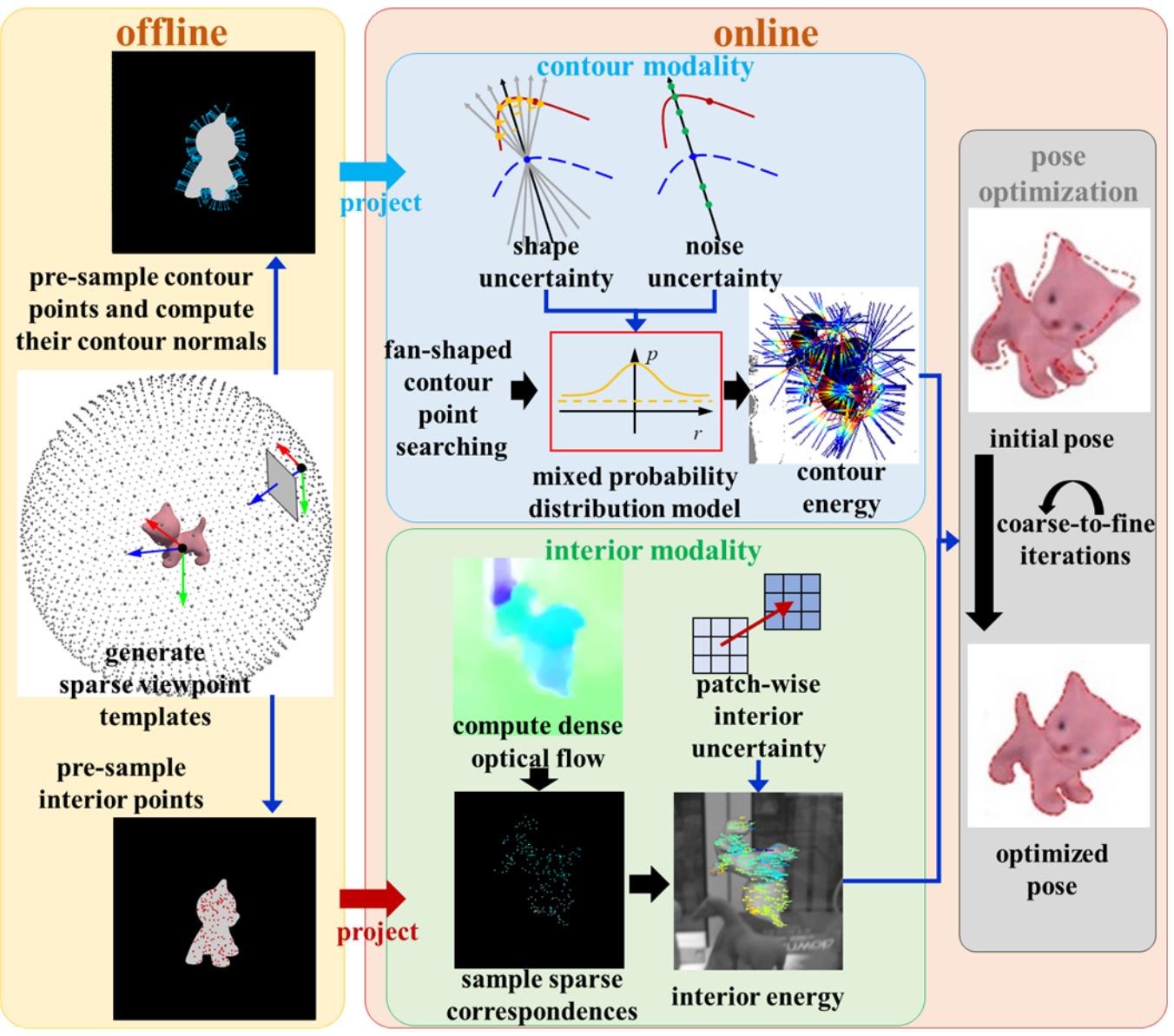

**Fig. 3.** Our 6DoF object pose tracking flowchart.

### *D. Deep Learning-based Methods*

The deep learning-based approaches predict the pose parameters directly by training neural networks. [42] utilized a lightweight CNN to predict object contour positions and uncertainties, achieving state-of-the-art performance in monocular pose tracking methods. Other method using "render-and-compare" strategy and iteratively refining pose [43] also achieved great performance. Although deep learning-based methods show potential, their dependence on additional GPU resources and significant time consumption render them impractical for resource-constrained industrial assembly applications. Furthermore, training networks for specific objects requires annotating 6DoF datasets [44], [45], which presents significant challenges and requires considerable time.

## III. Method

In this section, we describe our 6DoF object pose tracking method in detail. Fig. 3 illustrates the flowchart of our method. During offline stage, both contour and interior points are randomly sampled at pre-generated sparse viewpoint templates, and their 2D-3D correspondences are stored. In online tracking stage, we propose a fan-shaped search strategy to build point-to-distribution contour correspondences, taking into account the shape and noise uncertainties of local contours. Additionally, we utilize an efficient CPU-only strategy to establish sparse interior correspondences by using the DIS optical flow algorithm. Finally, we formulate a joint energy function that can simultaneously optimize contour and interior energies with a re-weighted least squares algorithm.

### *A. Preliminaries*

The purpose of 6DoF object pose tracking is to consecutively measure the 6DoF pose of a 3D object from image sequences, which can be represented by a 4 × 4 homogeneous matrix

$$\mathbf{T}_{CM} = \begin{bmatrix} \mathbf{R}_{CM} & \mathbf{t}_{CM} \\ \mathbf{0} & 1 \end{bmatrix} \in \mathbf{SE}(3). \tag{1}$$

If we define an object model point as $\mathbf{X}_M = [X_M, Y_M, Z_M]^{\mathrm{T}} \in \mathbf{R}^3$ and describe its camera coordinate as $\mathbf{X}_C = [X_C, Y_C, Z_C]^{\mathrm{T}} \in \mathbf{R}^3$, the transformation from a 3D model coordinate frame to its corresponding 2D image point can be written as

$$\mathbf{x} = \pi\left(\mathbf{K}\left(\mathbf{T}_{CM}\tilde{\mathbf{X}}_M\right)_{3\times1}\right), \tag{2}$$

where $\tilde{\mathbf{X}}_M$ is the homogeneous representation of $\mathbf{X}_M$, $\pi(\mathbf{X}) = [X/Z, Y/Z]^{\mathrm{T}}$ for 3D point $\mathbf{X}$ = [$X$, $Y$, $Z$], $\mathbf{K}$ is the known 3 × 3 camera intrinsic matrix. During optimization, pose variation is represented by a Lie algebra $\xi = [w_1, w_2, w_3, v_1, v_2, v_3]^{\mathrm{T}} \in \mathbf{R}^6$ and its corresponding transformation matrix can be computed by an exponential map

$$\Delta\mathbf{T}(\xi) = \exp(\hat{\xi}) \in \mathbf{SE}(3), \tag{3}$$

and with a previous pose represented by transformation matrix $\mathbf{T}_{CM}(t-1)$, the current object pose can be updated by

$$\mathbf{T}_{CM}(t) = \Delta\mathbf{T}(\xi)\mathbf{T}_{CM}(t-1). \tag{4}$$

### *B. Contour-based Tracking with Correspondence Uncertainty*

To compute the inter-frame motion $\Delta\mathbf{T}(\xi)$, contour-based methods must establish 3D-2D correspondences between 3D model points and object contour points. The challenge lies in accurately corresponding the projected contour point to the object contour point on the image. However, factors such as noise, cluttered backgrounds, and occlusion in industrial environments often cause visual ambiguities. For instance, a projected contour point may correspond to inaccurate image points, which can lead to tracking failure. In this section, we describe our contour correspondence searching approach which can effectively handle the unavoidable false contour points caused by these challenges.

*1) Pre-generated Sparse Viewpoint Templates:* In industrial assembly, tracking efficiency and minimizing GPU usage are crucial. To address this issue, we employ a strategy of pre-generating sparse virtual viewpoint templates, inspired by [32]. We uniformly sample 2562 viewpoints on a spherical surface centered at the object. A virtual camera oriented towards the object renders depth images. The z-axis direction vector of the virtual camera at the $j$-th viewpoint is denoted as $\mathbf{v}_j$. These depth images are then binarized to generate object masks. $N_{cnt}$ contour points along the object mask edges and $N_{in}$ interior points within the mask are subsequently sampled randomly. Using (5), we transform their 2D coordinates $\mathbf{x}_i^{\mathbf{v}_j}$ and contour normal vectors $\mathbf{n}_i^{\mathbf{v}_j}$ into the 3D model coordinates using depth $d_i^{\mathbf{v}_j}$, and store them as viewpoint templates.

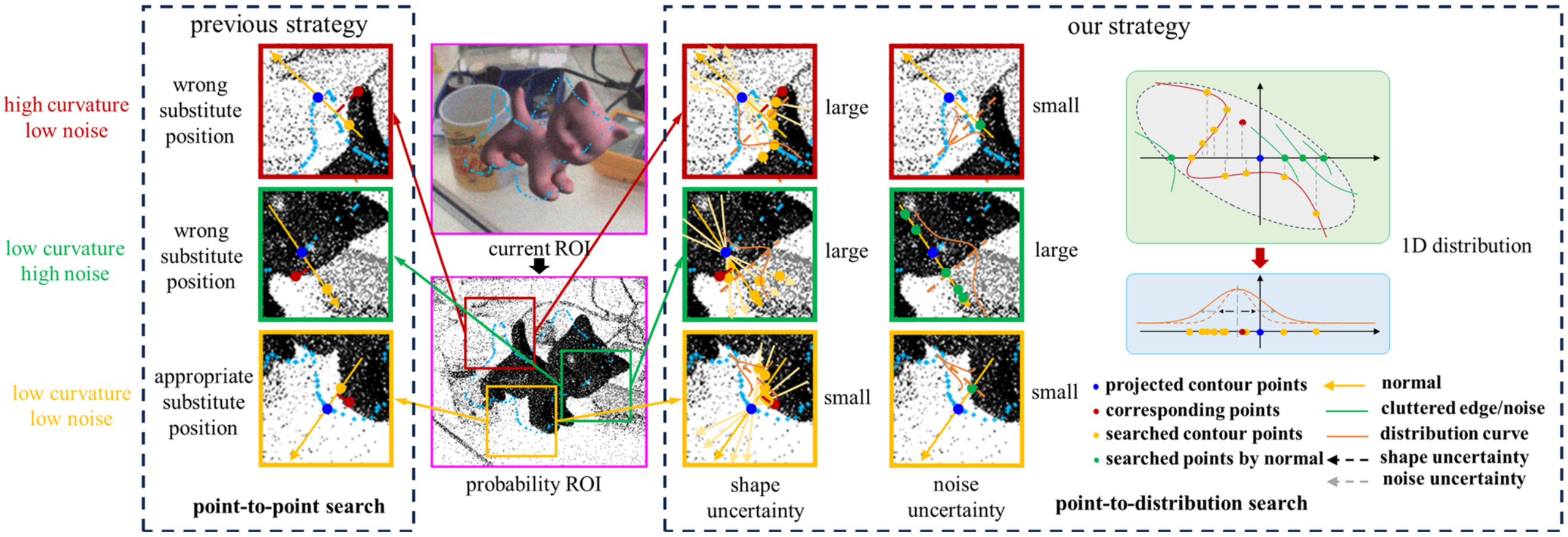

**Fig. 4.** Previous point-to-point correspondence search and our point-to-distribution correspondence search in different contour matching scenarios.

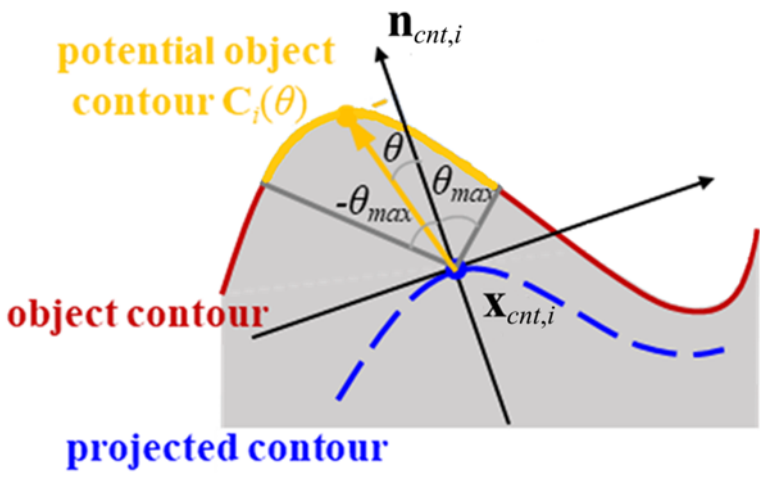

**Fig. 5.** Parametric equation representation of local contour.

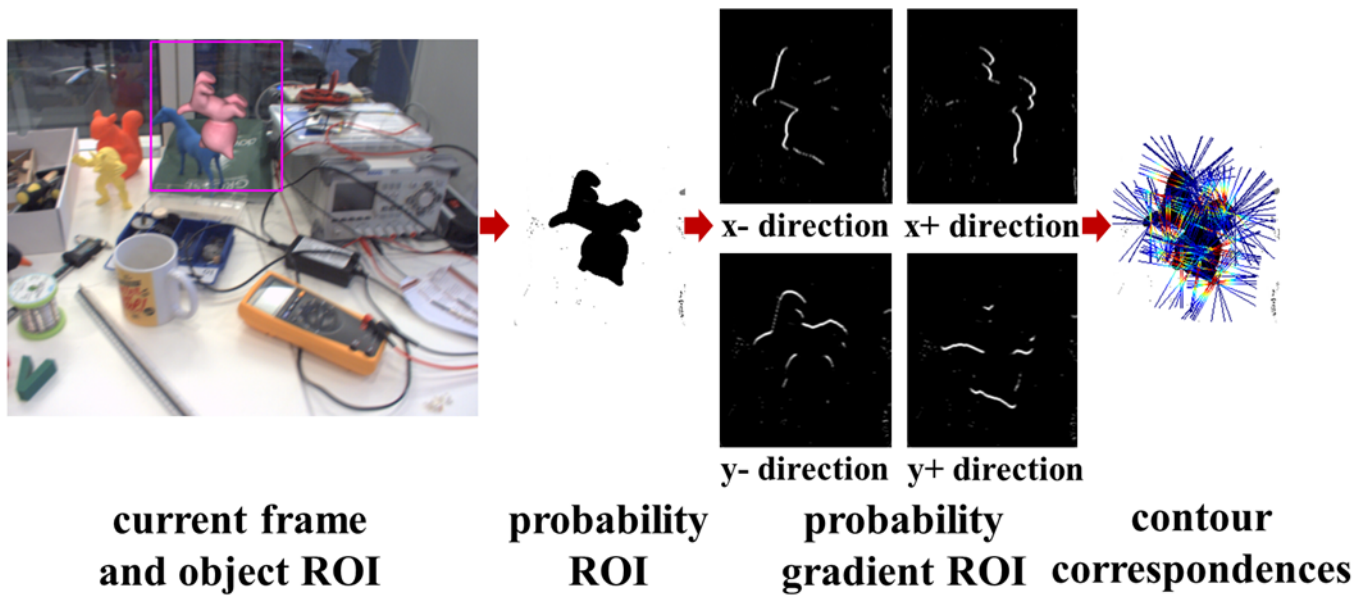

**Fig. 6.** Contour data association strategy. Contour correspondences are computed based on so-called probability gradient images. Then, fan-shaped search strategy and direction-gradient-based confidence are used to compute candidate contour points.

$$\tilde{\mathbf{X}}_i^{\mathbf{v}_j} = \mathbf{T}_{MC}^{\mathbf{v}_j}\pi^{-1}\left(\mathbf{x}_i^{\mathbf{v}_j}, d_i^{\mathbf{v}_j}\right),\ \mathbf{N}_i^{\mathbf{v}_j} = \mathbf{R}_{MC}^{\mathbf{v}_j}\left[\mathbf{n}_i^{\mathbf{v}_j \mathrm{T}}, 0\right]^{\mathrm{T}}. \quad (5)$$

*2) Contour Search Line:* During tracking, given a previous pose $\mathbf{T}_{CM}(t\text{-}1)$, we first calculate its closest virtual viewpoint $\mathbf{v}_j$. The 3D model points $\tilde{\mathbf{X}}_i^{\mathbf{v}_j}$ and normal vectors $\mathbf{N}_i^{\mathbf{v}_j}$ stored in its viewpoint template thus can be projected onto the current image $I(\mathbf{x})$. Therefore, we can obtain the projected contour of 3D model. Using the color probability, the object on the image can be segmented into a foreground region $\Omega_f$ and a background region $\Omega_b$ with the background color probability of pixel $\mathbf{x}$

$$P_b(\mathbf{x}) = p_b(I(\mathbf{x}))\big/\left(p_f(I(\mathbf{x})) + p_b(I(\mathbf{x}))\right), \quad (6)$$

where $p_f(I(\mathbf{x}))$ and $p_b(I(\mathbf{x}))$ denote current foreground and background color histograms, respectively.

For the $i$-th projected contour point $\mathbf{x}_{cnt,i}$, we define the $j$-th search line $L_{i,j}$ with $L_{src}$ as the length, $\mathbf{x}_{cnt,i}$ as the center, and $[l_{i,j,x}, l_{i,j,y}]^{\mathrm{T}}$ as the direction vector. For the $k$-th point $\mathbf{x}_{i,j,k}$ on the search line $L_{i,j}$, we calculate its direction gradient $g_{i,j,k}$ along the $[l_{i,j,x}, l_{i,j,y}]^{\mathrm{T}}$ on the probability image $P_b(\mathbf{x})$. We consider the points whose $g_{i,j,k} > 0$ as candidate contour points of this search line $L_{i,j}$. Although there are maybe more than one candidate points whose direction gradient are greater than 0, only the point with the highest direction gradient on this line is considered as the potential contour point.

*3) Contour Correspondence Determination:* In most contour-based methods, such as [28] and [30], the point searched along the normal of projected contour point $\mathbf{x}_{cnt,i}$ is treated as corresponding object contour point. However, this strategy is sensitive to contour curvature, large displacements, and noise, as shown in Fig. 4, which can lead to erroneous correspondences and degrade pose accuracy. To mitigate this problem, a reasonable method is to sample multiple local object contour points around the projected point $\mathbf{x}_{cnt,i}$ and consider the mean of their distribution as the most likely correspondence. This would help reduce the impact of abnormal sampling points. Therefore, we propose a fan-shaped search strategy to find potential object contour points within a fan region and project these points onto the normal of projected contour point to generate a 1D probability distribution.

As described in Fig. 5, given a projected contour point $\mathbf{x}_{cnt,i}$, we use a parametric equation $\mathbf{C}_i(\theta)$ to describe the potential object contour on the fan-shaped search region around it, where $\theta \in [-\theta_{max}, \theta_{max}]$ denotes the search angle range $A_{reg}$. These object contour points are then projected onto the normal $\mathbf{n}_{cnt,i}$ to form a 1D distribution. The mean of this distribution $\mu_i$ represents the expected position of object contour point along the normal, and the variance illustrates the shape complexity of this local contour which we refer to as shape uncertainty $\sigma_{shp,i}^2$. Since it is difficult to know the exact mathematical expression of $\mathbf{C}_i(\theta)$, We approximately estimate it by uniformly sampling an odd number of angles, $N_{sam}$

$$\mu_i = \frac{1}{2\theta_{max}} \int_{\theta=-\theta_{max}}^{\theta_{max}} \mathbf{n}_{cnt,i}^{\mathrm{T}} \mathbf{C}_i(\theta)\, d\theta \approx \frac{1}{N_{sam}} \sum_{j=1}^{N_{sam}} \mathbf{n}_{cnt,i}^{\mathrm{T}} \mathbf{C}_i(\theta_j). \quad (7)$$

To estimate the variance denoted as $\sigma_{shp,i}^2$, we impose the constraint $\sigma_{shp,i}^2 > 1$ to ensure stability in the pose optimization

$$\sigma_{shp,i}^{\ 2} = \max\left\{ \frac{1}{N_{sam}} \sum_{j=1}^{N_{sam}} \left( \mathbf{n}_{cnt,i}^{\mathrm{T}} \mathbf{C}_i(\theta_j) - \mu_i \right)^2, 1 \right\}. \quad (8)$$

Meanwhile, in scenarios with heavier additive noise and cluttered backgrounds that may obscure object contours and result in wrong contour correspondences during line search, we thus explicitly define a noise uncertainty $\sigma_{noi,i}^2$ and use it to scale $\sigma_{shp,i}^2$ to represent the correspondence uncertainty $\sigma_i^2 = \sigma_{shp,i}^2\sigma_{noi,i}^2$. Here, $\sigma_{noi,i}$ is defined as (9)

$$\sigma_{noi,i} = \sum_{g_{i,j,k}\in\{g_{i,j,k}\}} g_{i,j,k} \Big/ \max\{g_{i,j,k}\},\ j=(N_{sam}+1)/2, \quad (9)$$

which is the sum of direction gradients of all candidate contour points divided by the direction gradient of the potential contour point on the normal search line. Our contour data association strategy is depicted as Fig. 6.

*4) Contour Energy Derived from The Mixed Probability Model:* For a projected contour point $\mathbf{x}_{cnt,i}(\xi)$, the residual $r_{cnt,i}$ is defined as the 1D projection distance between $\mathbf{x}_{cnt,i}(\xi)$ and the mean $\mu_i$ described in (7). The probability density of $r_{cnt,i}$ is represented as $P(r_{cnt,i}|\xi)$

$$r_{cnt,i} = \mathbf{n}_{cnt,i}^{\mathrm{T}}\left(\mathbf{x}_{cnt,i}(\xi)-\mathbf{x}_{cnt,i}(\mathbf{0})\right)-\mu_i \sim P\left(r_{cnt,i}|\xi\right), \text{ with}$$
$$\mathbf{x}_{cnt,i}(\xi)=\pi\left(\mathbf{K}\left(\Delta\mathbf{T}(\xi)\mathbf{T}_{CM}\tilde{\mathbf{X}}_{cnt,i}^{v_j}\right)_{3\times1}\right). \quad (10)$$

Assuming that projected contour points are independent, the optimal pose $\xi$ can be obtained by minimizing the negative log of likelihood probability, thus the final contour energy function is defined as follows

$$E_{cnt} = -\ln\left(\prod_{i=1}^{N_{cnt}} P\left(r_{cnt,i}|\xi\right)\right) = -\sum_{i=1}^{N_{cnt}} \ln\left(P\left(r_{cnt,i}|\xi\right)\right). \quad (11)$$

Note that the form of energy function $E_{cnt}$ depends on the likelihood probability density function, so it is not an absolutely standard least squares problem. Inspired by [47], we modify the energy function $E_{cnt}$ by introducing a contour weight $w_{cnt,i}$. During each optimization iteration, $w_{cnt,i}$ can be seen as a constant. Thus, $E_{cnt}$ becomes a re-weighted least squares form that can be solved by Gauss-Newton method

$$E_{cnt} = \sum_{i=1}^{N_{cnt}} \frac{1}{2} w_{cnt,i} r_{cnt,i}^{\ 2}, \text{ with } w_{cnt,i} = -\frac{\partial \ln\left(P\left(r_{cnt,i}|\xi\right)\right)}{\partial r_{cnt,i}} \frac{1}{r_{cnt,i}}. \quad (12)$$

Typically, contour-based methods use a Gaussian distribution to describe the probability density $P(r_{cnt,i}|\xi)$. However, Gaussian distribution can only form an inverse variance weighting, making it susceptible to outliers. To address this issue, inspired by [48], we propose a mixed probability distribution that combines a Gaussian and a uniform distribution, expressed as follows

$$P\left(r_{cnt,i}|\xi\right) = a_{1,i} e^{-\frac{1}{2\sigma_i^2} r_{cnt,i}^{\ 2}} + a_{2,i}, \quad (13)$$

in (13), $a_{1,i}$ and $a_{2,i}$ are the weight coefficients corresponding to Gaussian and uniform distribution. Because $\ln(P(r_{cnt,i}|\xi))$ cannot be differentiated easily, we use the following form to approximate it

$$\ln\left(a_{1,i} e^{-\frac{1}{2\sigma_i^2} r_{cnt,i}^{\ 2}} + a_{2,i}\right) \approx b_{1,i} e^{-\frac{b_{2,i}}{\sigma_i^2} r_{cnt,i}^{\ 2}} + b_{3,i}, \quad (14)$$

here, $b_{1,i}$, $b_{2,i}$ and $b_{3,i}$ are parameters which are used to fit the log of likelihood probability. It is worth to mention that the two forms must be equal at three specific positions: $r_{cnt,i} = 0$, $r_{cnt,i} \rightarrow \infty$ and $r_{cnt,i} = \sigma_i$, to ensure the validity of (14). Thus $b_{1,i}$, $b_{2,i}$ and $b_{3,i}$ can be solved by using following formulas

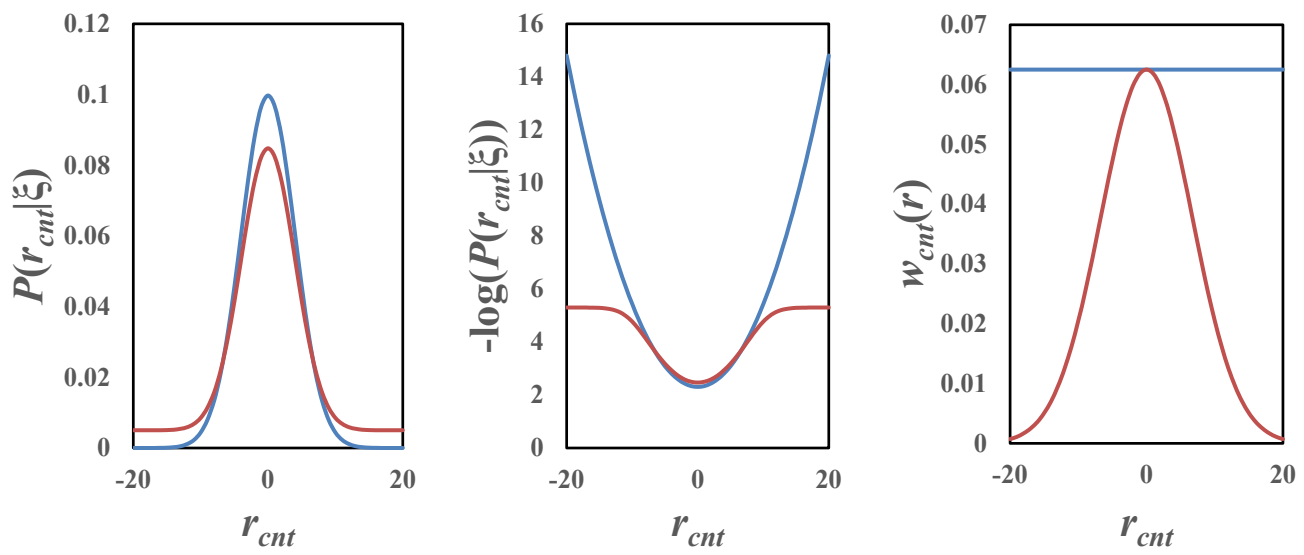

(a) (b) (c)

**Fig. 7.** Comparison between Gaussian probability model (blue curves) and mixed probability model (red curves) over different values of residual $r_{cnt}$. (a) Probability density function. (b) Negative log probability density. (c) Contour weight.

$$b_{1,i} = \ln\left(1+\frac{a_{1,i}}{a_{2,i}}\right),\ b_{2,i} = -\ln\left(\frac{\ln\left(1+\frac{a_{1,i}}{a_{2,i}}e^{-\frac{1}{2}}\right)}{\ln\left(1+\frac{a_{1,i}}{a_{2,i}}\right)}\right),\ b_{3,i} = \ln\left(a_{2,i}\right). \quad (15)$$

Substituting (14) into (12), we obtain the contour weight

$$w_{cnt,i} = \left(2b_{1,i}b_{2,i}/\sigma_i^2\right) e^{-\frac{b_{2,i}}{\sigma_i^2} r_{cnt,i}^{\ 2}}. \quad (16)$$

Assuming $a_{1,i} >> a_{2,i}$, meaning the weight coefficients of Gaussian is larger than that of uniform distribution, we can derive the following approximation

$$b_{1,i}b_{2,i} \approx -\ln\left(\frac{a_{1,i}}{a_{2,i}}\right)\ln\left(1+\ln\left(e^{-\frac{1}{2}}\right)\Big/\ln\left(\frac{a_{1,i}}{a_{2,i}}\right)\right) \approx \frac{1}{2}. \quad (17)$$

This approximation provides us a convenience to substitute $b_{1,i}b_{2,i}$ with a constant value 1/2 without having to compute them for all correspondences. To further simplify the contour weight $w_{cnt,i}$ described in (16), we again replace $b_{2,i}$ and $\sigma_i^2$ in exponential term with empirical values $b_2$ and $\sigma^2$ for all correspondences. Therefore, by defining $\beta = b_2 / \sigma^2$, we obtain the simplified formula of $w_{cnt,i}$

$$w_{cnt,i} = \left(1/\sigma_i^2\right) e^{-\beta r_{cnt,i}^{\ 2}} = \left(1/\left(\sigma_{shp,i}^{\ 2}\sigma_{noi,i}^{\ 2}\right)\right) e^{-\beta r_{cnt,i}^{\ 2}}. \quad (18)$$

Compared to the conventional Gaussian distribution assumption, modeling $P(r_{cnt,i}|\xi)$ as a mixture of a Gaussian and a uniform distribution can provide an additional exponential term $e^{-\beta r_{cnt,i}^{\ 2}}$. This modification helps depress outliers with large residuals, as illustrated in Fig. 7(c).

## *C. Efficient Interior Correspondence*

To further enhance the robustness of tracking rotationally symmetric objects and complement contour-based method to acquire improved reliability, we introduce additional sparse interior correspondences by computing dense optical flow in an efficient way. Since the 3D model points corresponding to the randomly sampled interior points within the object mask have been stored in the sparse viewpoint templates during the offline stage, we can use the closest virtual viewpoint to project the pre-stored 3D model interior points onto the image, obtaining the projected 2D interior points $\mathbf{x}_{in,i}$. For each 2D interior point $\mathbf{x}_{in,i}$, its correspondence $\mathbf{x}'_{in,i}$ can be obtained by sampling the dense optical flow field computed using DIS optical flow. To improve efficiency, we restrict the computation of dense optical flow to an ROI area.

After computing the interior correspondences, we aim to minimize re-projection errors between projected interior points $\mathbf{x}_{in,i}(\boldsymbol{\xi})$ and their correspondences $\mathbf{x}'_{in,i}$. Similar to the contour energy $E_{cnt}$, we express the interior energy $E_{in}$ as a re-weighted least squares problem by using (19) for effective optimization. Outliers can be depressed by adjusting $\gamma$

$$E_{in}=\sum_{i=1}^{N_{in}}\frac{1}{2}w_{in,i}r_{in,i}^{\ 2}\text{ , with } w_{in,i}=c_{in,i}e^{-\gamma r_{in,i}^{\ 2}}\text{ , and}$$

$$r_{in,i}=\left\|\mathbf{x}_{in,i}(\boldsymbol{\xi})-\mathbf{x}'_{in,i}\right\|_2\text{, with } \mathbf{x}_{in,i}(\boldsymbol{\xi})=\pi\left(\mathbf{K}\left(\Delta\mathbf{T}(\boldsymbol{\xi})\mathbf{T}_{CM}\tilde{\mathbf{X}}_{in,i}^{\mathbf{v}_j}\right)_{3\times1}\right),\quad (19)$$

here $c_{in,i}$ represents interior uncertainty and we defined it as

$$c_{in,i}=\max\left\{1-e_i^{\ 2},0\right\},\quad (20)$$

where $e_i$ is the measurement error of optical flow

$$e_i=e_{I,i}+e_{G,i}+e_{S,i}\ .\quad (21)$$

Instead of computing point-to-point error like [41], we take $s \times s$ patch around each interior point as a whole, and compute the error between the two corresponding patches to handle noise. Thus, we define the patch-wise intensity error $e_{I,i}$, gradient error $e_{G,i}$, and smoothness error $e_{S,i}$ as follows

$$e_{I,i}=\frac{1}{s^2}\left(\left|\sum_{j\in S_i}I\left(\mathbf{x}_j+\mathbf{u}_j\right)-\sum_{j\in S_i}I\left(\mathbf{x}_j\right)\right|\Big/\eta_I\right),$$

$$e_{G,i}=\frac{1}{s^2}\left(\left\|\sum_{j\in S_i}\nabla I\left(\mathbf{x}_j+\mathbf{u}_j\right)-\sum_{j\in S_i}\nabla I\left(\mathbf{x}_j\right)\right\|_2\Big/\eta_G\right),\quad (22)$$

$$e_{S,i}=\frac{1}{s^2}\left(\sum_{j\in S_i}\sqrt{\left\|\nabla\mathbf{u}_{j,x}\right\|_2^{\ 2}+\left\|\nabla\mathbf{u}_{j,y}\right\|_2^{\ 2}}\Big/\eta_S\right),$$

where $\eta_I$ = 40, $\eta_G$ = 40 and $\eta_S$ = 80 are regularization parameters, $S_i$ denotes pixels set within the $i$-th patch, $\mathbf{u}_j$ is the optical flow at $j$-th pixel $\mathbf{x}_j$, patch size $s$ = 3.

### *D. Joint Pose Optimization*

Given the contour energy $E_{cnt}$ and interior energy $E_{in}$, we establish a joint energy function. Here, $\lambda$ serves as a variable weight, allowing for the adjustment of the relative contribution of $E_{cnt}$ and $E_{in}$

$$E=\lambda E_{cnt}+(1-\lambda)E_{in}\ .\quad (23)$$

Considering that optical flow can estimate large displacement of objects, but can hardly achieve pixel-level accuracy, and contour-based method may easily fall in local minima but can reach pixel-level accuracy for small displacement. Therefore, we leverage the strengths of both modalities. As the number of optimization iterations increases, we adjust $\lambda$ from small to large, thus two modalities can better complement each other.

For each residual of contour and interior correspondence, the Jacobian $\mathbf{J}_{cnt,i}$ and $\mathbf{J}_{in,i}$ are computed by using the chain rule

$$\mathbf{J}_{cnt,i}=\frac{\partial r_{cnt,i}}{\partial\mathbf{x}_{cnt,i}}\frac{\partial\mathbf{x}_{cnt,i}}{\partial\mathbf{X}_{C,i}}\frac{\partial\mathbf{X}_{C,i}}{\partial\boldsymbol{\xi}}\ ,\ \mathbf{J}_{in,i}=\frac{\partial r_{in,i}}{\partial\mathbf{x}_{in,i}}\frac{\partial\mathbf{x}_{in,i}}{\partial\mathbf{X}_{C,i}}\frac{\partial\mathbf{X}_{C,i}}{\partial\boldsymbol{\xi}}\ .\quad (24)$$

Thus, the gradient vector of energy function is computed by

$$\mathbf{g}=\lambda\sum_{i=1}^{N_{cnt}}w_{cnt,i}\mathbf{J}_{cnt,i}^{\ \mathrm{T}}r_{cnt,i}+(1-\lambda)\sum_{i=1}^{N_{in}}w_{in,i}\mathbf{J}_{in,i}^{\ \mathrm{T}}r_{in,i}\ ,\quad (25)$$

**Algorithm 1** Joint pose optimization

**Input:** current RGB image $I(\mathrm{x})$, current ROI rectangular;
**Output:** current 6DoF object pose $\mathbf{T}_{CM}$;

```
Compute probability gradients, grayscale, and optical flow of the ROI;
Search for the closest view index j ← argmax_j (v_j^T R_MC t_CM);
Compute x'_in,i and c_in,i for each interior correspondence using (20);
for k ← {1, 2, 2, 4} do
  A_reg ← {A_reg}, L_src ← {L_src}, σ ← {σ}, γ ← {γ}, and λ ← {λ};
  for m ← {1, …, k} do
    Search for the closest view index j ← argmax_j (v_j^T R_MC t_CM);
    Compute μ_i and σ_i for each contour correspondence using (7), (8), (9);
    for n ← {1, 2, 3} do
      Update weights w_cnt,i ← (1/σ_i^2) e^{-(b_2/σ^2) r_cnt,i^2}, w_in,i ← c_in,i e^{-γ r_in,i^2};
      Compute gradient vector g and Hessian H using (25) and (26);
      Update current pose T_CM using (27) and (4);
    end for
  end for
end for
Update color histograms, ROI rectangular and grayscale of the ROI;
```

and Hessian matrix can be approximately calculated by

$$\mathbf{H}\approx\lambda\sum_{i=1}^{N_{cnt}}w_{cnt,i}\mathbf{J}_{cnt,i}^{\ \mathrm{T}}\mathbf{J}_{cnt,i}+(1-\lambda)\sum_{i=1}^{N_{in}}w_{in,i}\mathbf{J}_{in,i}^{\ \mathrm{T}}\mathbf{J}_{in,i}\ .\quad (26)$$

Thus, for each iteration, the pose variation can be solved by regularized Gauss-Newton method

$$\boldsymbol{\xi}=\left(-\mathbf{H}+\begin{bmatrix}\lambda_r\mathbf{I}_3 & 0\\ 0 & \lambda_t\mathbf{I}_3\end{bmatrix}\right)^{-1}\mathbf{g}\ ,\quad (27)$$

where $\mathbf{H}$ is the Hessian matrix and $\mathbf{g}$ is the gradient vector of $E$, regularization parameters $\lambda_r$ = 5000, $\lambda_t$ = 500000. The joint pose optimization process is illustrated in Algorithm 1.

## IV. Implementation

In experiments, we expand the bounding box of object by 40 pixels as ROI area, which gives consideration of computational efficiency and displacement range of object. Sobel operator is used to approximately compute gradient of probability image. The number of contour sampling points $N_{cnt}$ is equal to the number of interior sampling points $N_{in}$, we set them to 200. To improve the efficiency and accuracy of our method, the coarse-to-fine fan-shaped search and optimization strategies are employed. Before tracking, we uniformly define 36 directions to cover 360° and precompute the cosine values for the angles between each pair of directions. These predefined directions are used to approximate the sampling directions during the tracking stage.

During tracking, for the $i$-th projected contour point $\mathbf{x}_{con,i}$, we initially approximate its normal $\mathbf{n}_{cnt,i}$ to the closest predefined direction, and search starts with the widest search range $A_{reg}$ and the longest search lines length $L_{src}$ to handle significant displacements. Then, the length of the search lines and the search range are progressively decreased to facilitate optimization convergence.

In our experiments, our defined sampling interval angle $\Delta A_{int}$ is 10°, search region angle in each step $\{A_{reg}\}$ is {60°, 40°, 20°, 0°}, and the length of search lines in each step $\{L_{src}\}$ is {73 pixels, 43 pixels, 23 pixels, 13 pixels}.

TABLE I
EXPERIMENTAL RESULTS ON RBOT DATASET. WE COMPARE WITH STATE-OF-THE-ART OPTIMIZATION-BASED 6DOF POSE TRACKING METHODS. THE BEST RESULTS ARE BOLD, THE SECOND-BEST RESULTS ARE UNDERLINED, AND THE ROTATIONALLY SYMMETRIC OBJECTS ARE COLORED RED.

| Variant | Method | Ape | Soda | Vise | Soup | Camera | Can | Cat | Clown | Cube | Driller | Duck | Egg Box | Glue | Iron | Candy | Lamp | Phone | Squirrel | Average |
|---|---|---|---|---|---|---|---|---|---|---|---|---|---|---|---|---|---|---|---|---|
| Regular | TPAMI19 [25] | 85.0 | 39.0 | 98.9 | 82.4 | 79.7 | 87.6 | 95.9 | 93.3 | 78.1 | 93.0 | 86.8 | 74.6 | 38.9 | 81.0 | 46.8 | 97.5 | 80.7 | 99.4 | 79.9 |
| | TIP20 [26] | 88.8 | 41.3 | 94.0 | 85.9 | 86.9 | 89.0 | 98.5 | 93.7 | 83.1 | 87.3 | 86.2 | 78.5 | 58.6 | 86.3 | 57.9 | 91.7 | 85.0 | 96.2 | 82.7 |
| | TIE21 [39] | 93.7 | 39.3 | 98.4 | 91.6 | 84.6 | 89.2 | 97.9 | 95.9 | 86.3 | 95.1 | 93.4 | 77.7 | 61.5 | 87.8 | 65.0 | 95.2 | 85.7 | 99.8 | 85.5 |
| | JCST21 [36] | 92.8 | 42.6 | 96.8 | 87.5 | 90.7 | 86.2 | 99.0 | 96.9 | 86.8 | 94.6 | 90.4 | 87.0 | 57.6 | 88.7 | 59.9 | 96.5 | 90.6 | 99.5 | 85.8 |
| | TCSVT21 [35] | 93.0 | 55.2 | 99.3 | 85.4 | 96.1 | 93.9 | 98.0 | 95.6 | 79.5 | 98.2 | 89.7 | 89.1 | 66.5 | 91.3 | 60.6 | 98.6 | 95.6 | 99.6 | 88.1 |
| | TVCG22 [27] | 94.6 | 49.4 | 99.5 | 91.0 | 93.7 | 96.0 | 97.8 | 96.6 | 90.2 | 98.2 | 93.4 | 90.3 | 64.4 | 94.0 | 79.0 | **98.8** | 92.9 | 99.8 | 89.9 |
| | IJCV22 [28] | 98.8 | 65.1 | 99.6 | 96.0 | 98.0 | 96.5 | **100.0** | 98.4 | 94.1 | 96.9 | 98.0 | 95.3 | 79.3 | 96.0 | 90.3 | 97.4 | 96.2 | 99.8 | 94.2 |
| | ECCV22 [30] | **99.8** | 67.1 | **100.0** | **97.8** | 97.3 | 93.7 | **100.0** | 99.4 | 97.4 | 97.6 | **99.3** | 96.9 | 84.7 | **97.7** | 93.4 | 96.7 | 95.4 | **100.0** | 95.2 |
| | TIE23 [29] | - | - | - | - | - | - | - | - | - | - | - | - | - | - | - | - | - | - | 94.6 |
| | AEI24 [37] | 97.7 | 54.2 | 99.6 | 96.9 | 99.1 | **98.7** | **100.0** | **99.5** | 95.6 | 98.9 | 96.9 | 96.9 | 83.0 | 97.1 | 95.3 | 97.6 | 97.6 | 99.8 | 94.7 |
| | Ours (C) | 98.2 | 64.9 | 99.4 | 96.8 | 97.8 | 96.6 | 99.7 | 97.9 | 94.2 | 97.1 | 96.6 | 95.5 | 83.7 | 95.1 | 92.3 | 97.0 | 96.2 | 99.8 | 94.4 |
| | Ours | 99.0 | **85.0** | 99.6 | 97.5 | **99.6** | 98.1 | **100.0** | 98.8 | **97.5** | **98.6** | 98.9 | **98.5** | **87.4** | 97.2 | **95.4** | 98.1 | **98.7** | **100.0** | **97.1** |
| Dynamic Light | TPAMI19 [25] | 84.9 | 42.0 | 99.0 | 81.3 | 84.3 | 88.9 | 95.6 | 92.5 | 77.5 | 94.6 | 86.4 | 77.3 | 52.9 | 77.9 | 47.9 | 96.9 | 81.7 | 99.3 | 81.2 |
| | TIP20 [26] | 89.7 | 40.2 | 92.7 | 86.5 | 86.6 | 89.2 | 98.3 | 93.9 | 81.8 | 88.4 | 83.9 | 76.8 | 55.3 | 79.3 | 54.7 | 88.7 | 81.0 | 95.8 | 81.3 |
| | TIE21 [39] | 93.5 | 38.2 | 98.4 | 88.8 | 87.0 | 88.5 | 98.1 | 94.4 | 85.1 | 95.1 | 92.7 | 76.1 | 58.1 | 79.6 | 62.1 | 93.2 | 84.7 | 99.6 | 84.1 |
| | JCST21 [36] | 93.5 | 43.1 | 96.6 | 88.5 | 92.8 | 86.0 | 99.6 | 95.5 | 85.7 | 96.8 | 91.1 | 90.2 | 68.4 | 86.8 | 59.7 | 96.1 | 91.5 | 99.2 | 86.7 |
| | TCSVT21 [35] | 93.8 | 55.9 | 99.6 | 85.6 | 97.7 | 93.7 | 97.7 | 96.5 | 78.3 | 98.6 | 91.0 | 91.6 | 72.1 | 90.7 | 63.0 | **98.9** | 94.4 | **100.0** | 88.8 |
| | TVCG22 [27] | 94.3 | 48.3 | 99.5 | 90.1 | 94.6 | 96.1 | 97.9 | 97.3 | 90.9 | **99.1** | 92.9 | 91.5 | 72.6 | 94.7 | 80.0 | 98.3 | 95.2 | 99.8 | 90.7 |
| | IJCV22 [28] | 98.2 | 65.2 | 99.2 | 95.6 | 97.5 | 98.1 | **100.0** | 98.5 | 94.2 | 97.5 | 97.9 | 96.9 | 86.1 | 95.2 | 89.3 | 97.0 | 95.9 | 99.9 | 94.6 |
| | ECCV22 [30] | **100.0** | 64.5 | **99.8** | **97.9** | 97.9 | 94.0 | **100.0** | **99.5** | 97.0 | 98.8 | **99.3** | 97.6 | 87.5 | 97.4 | 92.4 | 97.1 | 96.4 | **100.0** | 95.4 |
| | TIE23 [29] | - | - | - | - | - | - | - | - | - | - | - | - | - | - | - | - | - | - | 95.2 |
| | AEI24 [37] | 98.5 | 49.7 | **99.8** | 97.1 | 98.6 | **98.8** | **100.0** | 99.2 | **97.1** | 98.9 | 97.9 | 98.4 | 87.8 | **97.9** | 94.4 | 97.9 | 97.8 | **100.0** | 95.0 |
| | Ours (C) | 98.8 | 63.1 | 99.6 | 96.4 | 97.7 | 97.8 | 99.5 | 98.2 | 94.7 | 97.4 | 96.8 | 97.1 | 88.5 | 94.8 | 92.3 | 97.2 | 96.4 | 99.6 | 94.8 |
| | Ours | 99.0 | **80.5** | **99.8** | 97.2 | **99.7** | 98.4 | 99.7 | 99.3 | **97.1** | **99.1** | 98.2 | **99.0** | **89.0** | 97.1 | **96.2** | **98.9** | **98.8** | **100.0** | **97.0** |
| Noise + Dynamic Light | TPAMI19 [25] | 77.5 | 44.5 | 91.5 | 82.9 | 51.7 | 38.4 | 95.1 | 69.2 | 24.4 | 64.3 | 88.5 | 11.2 | 2.9 | 46.7 | 32.7 | 57.3 | 44.1 | 96.6 | 56.6 |
| | TIP20 [26] | 79.3 | 35.2 | 82.6 | 86.2 | 65.1 | 56.9 | 96.9 | 67.0 | 37.5 | 75.2 | 85.4 | 35.2 | 18.9 | 63.7 | 35.4 | 64.6 | 66.3 | 93.2 | 63.6 |
| | TIE21 [39] | 84.7 | 33.0 | 88.8 | 89.5 | 56.4 | 50.1 | 94.1 | 66.5 | 32.3 | 79.6 | 94.2 | 29.6 | 19.9 | 63.4 | 40.3 | 61.6 | 62.4 | 96.9 | 63.5 |
| | JCST21 [36] | 89.1 | 44.0 | 91.6 | 89.4 | 75.2 | 62.3 | 98.6 | 77.3 | 41.2 | 81.5 | 91.6 | 54.5 | 31.8 | 65.0 | 46.0 | 78.5 | 69.6 | 97.6 | 71.4 |
| | TCSVT21 [35] | 92.5 | 56.2 | **98.0** | 85.1 | 91.7 | 79.0 | 97.7 | 86.2 | 40.1 | 96.6 | 90.8 | 70.2 | 50.9 | 84.3 | 49.9 | 91.2 | 89.4 | 99.4 | 80.5 |
| | TVCG22 [27] | 91.0 | 49.1 | 95.6 | 91.0 | 76.3 | 54.1 | 97.1 | 73.7 | 27.3 | 92.8 | 95.3 | 30.2 | 7.8 | 73.9 | 56.8 | 71.4 | 70.8 | 98.7 | 69.6 |
| | IJCV22 [28] | 96.9 | 61.9 | 95.4 | 95.7 | 84.5 | 73.9 | **99.9** | 90.3 | 62.2 | 87.8 | 97.6 | 62.2 | 43.4 | 84.3 | 78.2 | 73.3 | 83.1 | 99.7 | 81.7 |
| | ECCV22 [30] | **99.3** | 62.0 | 95.8 | 97.7 | 90.4 | 68.6 | **99.9** | 91.3 | 54.2 | 95.4 | **99.0** | 64.8 | 51.6 | 89.2 | 75.2 | 74.7 | 87.6 | **100.0** | 83.2 |
| | TIE23 [29] | - | - | - | - | - | - | - | - | - | - | - | - | - | - | - | - | - | - | 83.1 |
| | AEI24 [37] | 98.0 | 50.3 | 95.1 | **98.5** | 93.8 | 83.2 | **99.9** | 89.9 | 65.2 | 94.7 | 97.2 | 79.5 | 59.2 | 89.6 | 86.3 | 77.7 | 90.0 | 99.9 | 86.0 |
| | Ours (C) | 98.4 | 60.9 | 95.9 | 95.4 | 91.3 | 80.0 | 98.9 | 87.7 | 65.3 | 94.8 | 96.6 | 71.4 | 54.8 | 88.3 | 79.4 | 83.2 | 89.4 | 99.6 | 85.1 |
| | Ours | 98.7 | **78.1** | 97.5 | 97.1 | **96.7** | **90.0** | 99.0 | **95.1** | **82.2** | **98.3** | 97.7 | **86.1** | **66.5** | **92.7** | **91.6** | **93.4** | **95.1** | **100** | **92.0** |
| Occlusion + Dynamic Light | TPAMI19 [25] | 80.0 | 42.7 | 91.8 | 73.5 | 76.1 | 81.7 | 89.8 | 82.6 | 68.7 | 86.7 | 80.5 | 67.0 | 46.6 | 64.0 | 43.6 | 88.8 | 68.6 | 86.2 | 73.3 |
| | TIP20 [26] | 83.9 | 38.1 | 92.4 | 81.5 | 81.3 | 85.5 | 97.5 | 88.9 | 76.1 | 87.5 | 81.7 | 72.7 | 52.5 | 77.2 | 53.9 | 88.5 | 79.3 | 92.5 | 78.4 |
| | TIE21 [39] | 87.1 | 36.7 | 91.7 | 78.8 | 79.2 | 82.5 | 92.8 | 86.1 | 78.0 | 90.2 | 83.4 | 72.0 | 52.3 | 72.8 | 55.9 | 86.9 | 77.8 | 93.0 | 77.6 |
| | JCST21 [36] | 89.3 | 43.3 | 92.2 | 83.1 | 84.1 | 79.0 | 94.5 | 88.6 | 76.2 | 90.4 | 87.0 | 80.7 | 61.6 | 75.3 | 53.1 | 91.1 | 81.9 | 93.4 | 80.3 |
| | TCSVT21 [35] | 91.3 | 56.7 | 97.8 | 82.0 | 92.8 | 89.9 | 96.6 | 92.2 | 71.8 | 97.0 | 85.0 | 84.6 | 66.9 | 87.7 | 56.1 | 95.1 | 89.8 | 98.2 | 85.1 |
| | TVCG22 [27] | 92.5 | 51.5 | 99.2 | 90.7 | 92.1 | 92.2 | 97.7 | 94.2 | 89.8 | **98.4** | 91.3 | 90.7 | 66.3 | 91.7 | 75.3 | 95.9 | 92.1 | 99.0 | 88.9 |
| | IJCV22 [28] | 96.5 | 66.8 | 99.0 | 95.8 | 95.0 | 95.9 | **100.0** | 97.6 | 92.2 | 96.6 | 95.0 | 94.4 | 79.0 | 94.7 | 89.8 | 95.7 | 93.6 | 99.6 | 93.2 |
| | ECCV22 [30] | **98.7** | 68.4 | **99.9** | **97.5** | 98.3 | 93.0 | 99.9 | **99.4** | 95.1 | 97.9 | **99.1** | 96.9 | 85.5 | 97.0 | 90.3 | 96.3 | 95.1 | **100.0** | 94.9 |
| | TIE23 [29] | - | - | - | - | - | - | - | - | - | - | - | - | - | - | - | - | - | - | 93.6 |
| | AEI24 [37] | 96.7 | 53.8 | 99.4 | 96.0 | 98.1 | **98.0** | 99.7 | 98.8 | **95.7** | 98.0 | 95.0 | 97.8 | 84.8 | **97.1** | 92.6 | 96.0 | 96.3 | 99.9 | 94.1 |
| | Ours (C) | 96.2 | 62.6 | 98.5 | 94.4 | 96.1 | 96.2 | 99.1 | 96.9 | 92.3 | 97.0 | 94.0 | 95.0 | 84.1 | 93.7 | 89.9 | 95.6 | 94.4 | 98.9 | 93.1 |
| | Ours | 97.3 | **78.1** | 99.4 | 96.4 | **99.2** | 96.6 | 99.4 | 97.8 | **95.7** | **98.4** | 96.1 | **98.1** | **86.8** | 96.6 | **93.2** | **96.6** | **96.8** | 99.4 | **95.6** |

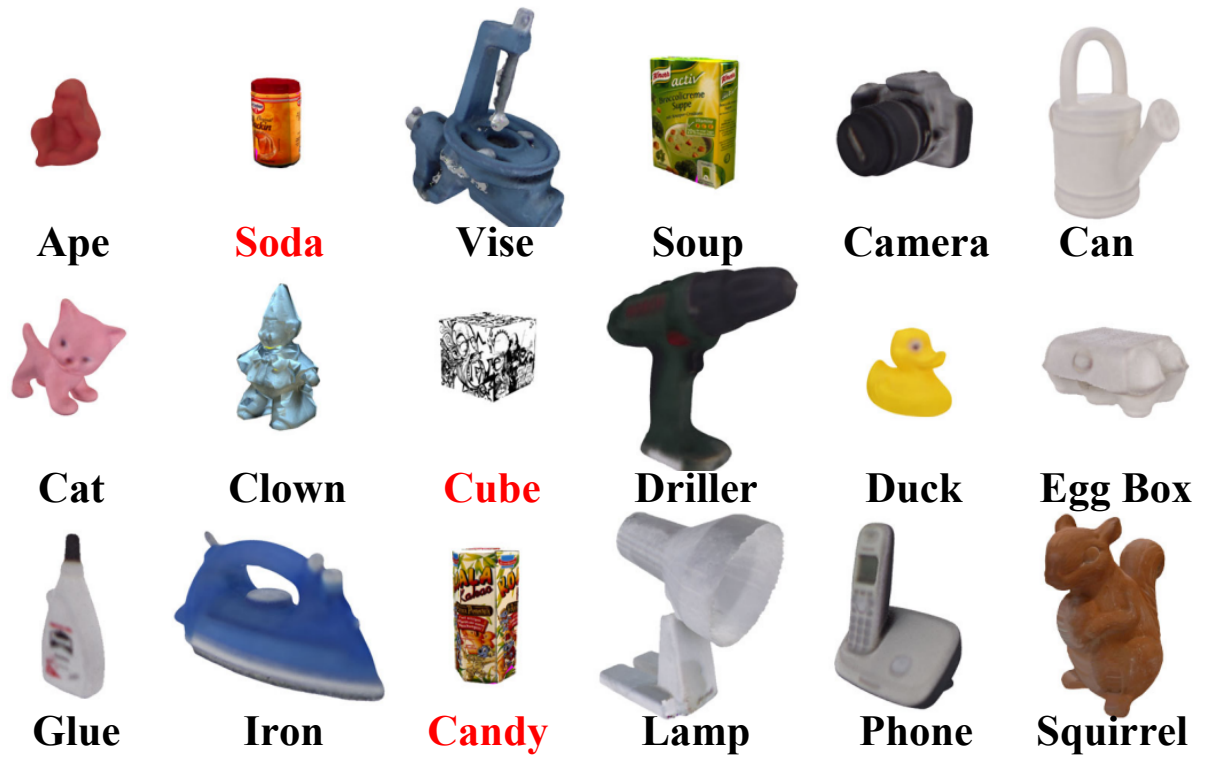

**Fig. 8.** 3D objects used in RBOT dataset [25]. Soda, Cube and Candy are rotationally symmetric objects.

For optimization strategy, $b_2$ is set to 0.2. Since the values of $\sigma$, $\gamma$ and $\lambda$ all depend on different coarse-to-fine optimization steps, we use sets $\{\sigma\}$, $\{\gamma\}$ and $\{\lambda\}$ to represent them. In our experiments, $\{\sigma\}$ is {8 pixels, 4 pixels, 2 pixels, 1 pixel}, $\{\gamma\}$ is {0.005, 0.15, 1.0, 2.0}, $\{\lambda\}$ is {0.4, 0.6, 0.8, 0.9}. Furthermore, we discard contour correspondences whose variance $\sigma_i^2$ is larger than 600.

## V. EXPERIMENT

The performances of our methods are evaluated and compared with state-of-the-art monocular 6DoF object tracking methods on the public datasets RBOT [25], OPT [49] and BCOT [50]. We also test our method in a real-world AR assembly guidance application. For simplicity, we refer to our contour-based tracker as "Ours (C)" and our multi-feature-based tracker as "Ours". All experiments run on a laptop with a 2.3 GHz Intel Core i7-12700H CPU and an NVIDIA GeForce RTX3060 GPU. The GPU is used only for offline generation of sparse viewpoint templates, while our 6DoF pose tracking algorithm runs on the CPU. C++ is used to implement our methods.

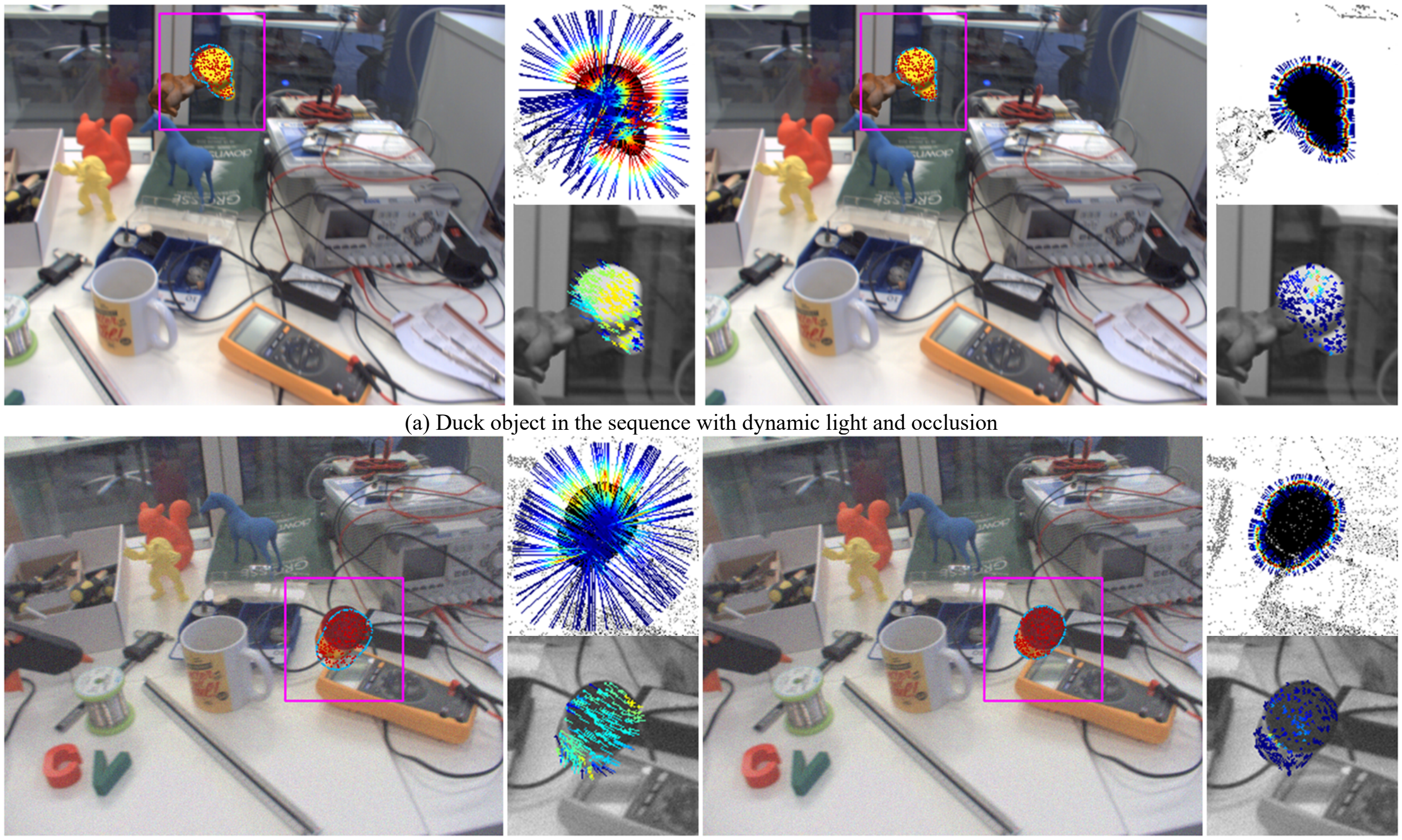

(a) Duck object in the sequence with dynamic light and occlusion

(b) Soda object which is rotationally symmetric in the sequence with noise

**Fig. 9.** Tracking examples. The first column and the third column show the projected sampling points before and after optimization, blue and red points represent the projected contour and interior points respectively. The second column and the fourth column show the coarsest and finest optimization step of our contour modality and interior modality. For contour modality, we visualize the weights of all points on all 1D distributions (optimization uses only the weights of the center points) to facilitate the demonstration of line search from coarse to fine. For interior modality, we visualize only the weights of the optical flow residuals to maintain a clean presentation. Warmer color indicates higher weight for each correspondence.

### *A. Evaluation on RBOT Dataset*

The RBOT dataset [25] is a semi-synthetic monocular 6DoF tracking dataset comprising 18 objects with diverse shapes and textures, as shown in Fig. 8. Each object has four image sequences, corresponding to regular case, case of dynamic light, case of dynamic light with Gaussian noise and case of dynamic light with occlusion. Each sequence contains 1001 monocular images of 640 × 512 pixels resolution, for each frame ground-truth rotation matrix $\mathbf{R}_{gt}$ and translation vector $\mathbf{t}_{gt}$ are labeled. In our experiments, we use the metric defined by [25], and respectively compute translation error and rotation error for each frame by

$$e_t = \left\| \mathbf{t} - \mathbf{t}_{gt} \right\|_2,\ e_R = \cos^{-1}\left( \left( trace\left( \mathbf{R}^{\mathrm{T}} \mathbf{R}_{gt} \right) - 1 \right) \big/ 2 \right), \quad (28)$$

pose tracking is considered successful only when $e_t < 5$cm and $e_R < 5°$, otherwise it is considered failed, and pose will be recovered by ground-truth.

Table I compares the tracking success rates of our method with those of state-of-the-art monocular methods, including [25]-[30], [35]-[37], and [39]. For contour-based methods, the most competitive methods are [28], a highly efficient sparse region-based approach, and [30], which combines local and non-local optimization to avoid local minima. When comparing the tracking success rates in regular, dynamic light and dynamic light with occlusion variants, "Ours (C)" has almost the same outstanding performance as [28] but is slightly inferior to [30]. In the dynamic light with noise scenario, "Ours (C)" achieves tracking success rates that are significantly 3.4% and 1.9% higher than those of [28] and [30], respectively. This demonstrates that our method is more robust in handling noises. This performance is expected, because [28] and [30] only use one direction normal search line to determine correspondence, which makes them more susceptible to errors in noisy frames. Our method addresses this problem by employing a fan-shaped search strategy to establish a 1D distribution of potential contour correspondences. The mixed probability distribution model used in our method contributes to a more robust energy function and effectively decrease outliers.

As shown in Table I, "Ours" significantly enhances tracking success rates compared to "Ours (C)", outperforming nearly all monocular tracking techniques across most sequences. Considering the average success rate for all sequences, "Ours" performs about 4.5% higher than that of [28], 3.2% higher than [30]. Notably, our method achieves a 17.7% improvement compared to [39], which also utilizes both contour and interior information. Though [37] fused the region-based method [28], non-local optimization [30], and additional edge information, we still achieve an average improvement of 2.9%.

When tracking the rotationally symmetric objects, such as Soda, Cube and Candy, "Ours" demonstrates the highest average tracking success rates of 80.4%, 93.0% and 94.1%. This is mainly attributed to additional interior correspondences that restrict the arbitrariness of out-of-plane rotations. Thus, the optimized pose direction is constrained, and incorrect correspondences are suppressed by our effective contour weight and interior weight. Also, our coarse-to-fine optimization strategy works to make two modalities complement each other.

Fig. 9 shows visualization results of "Ours" on RBOT dataset. During the coarsest correspondence search step, if the curvature around the projected contour point is large, or if there are significant noises, cluttered backgrounds, or occlusions, the corresponding 1D distributions exhibit higher uncertainty and are thus assigned lower weights. In the finest search step, shorter search lines, along with the smaller residual and uncertainty, enable the projected contour points to converge more effectively to object contours. Even erroneous correspondences caused by noise and occlusion can be reduced. As shown in Fig. 9(a), contour correspondences in the occluded area are assigned low weights, minimizing their impact on optimization. Similarly, in Fig. 9(b), contour correspondences in noisy regions also receive low weights. Furthermore, our interior weight effectively handles erroneous interior correspondences.

### B. Evaluation on OPT Dataset

The OPT dataset [49] is a real-world dataset that consists of 6 objects and 552 image sequences, each with a resolution of 1920 × 1080 pixels, captured under various lighting and motion conditions. Fig.10 illustrates partial examples from this dataset. To evaluate our methods, we follow the metric in [49], and compute the average vertex error of each frame

$$e_T = \frac{1}{n}\sum_{i=1}^{n}\left\|\left(\mathbf{T}\mathbf{X}_i - \mathbf{T}_{gt}\mathbf{X}_i\right)_{3\times1}\right\|_2 , \quad (29)$$

where $\mathbf{T}_{gt}$ is the ground truth pose and $\mathbf{X}_i$ is the $i$-th vertex of the 3D model of object. If $e_T < k_e d_m$, tracking is considered successful. Here $d_m$ is the largest distance between vertices and $k_e$ is a threshold varying from 0 to 0.2. Thus, tracking quality for each object is computed by an AUC (area under curve) score ranging from 0 to 20. During tracking process, initial pose of the first frame is given. But unlike RBOT dataset, if tracking fails in the OPT dataset, pose will not be recovered.

Experimental results on the OPT dataset are summarized in Table II. We compared our method with the state-of-the-art monocular 6DoF tracking methods [23], [25], [26], [28], [30], and [36], all of which only utilize monocular RGB images. Notably, when Compared with the most competitive contour-based method [28], "Ours (C)" achieves higher AUC scores for four out of six objects. And when it comes to the average AUC scores, "Ours (C)" has a 0.39 advantage over [28], and also outperforms other monocular methods, demonstrating its ability to acquire high robustness and accuracy in real cases. Also, we are surprisingly to find that for Jet object, "Ours (C)" gains outstanding AUC score 17.23, highlighting its capability to track objects with sharp contours.

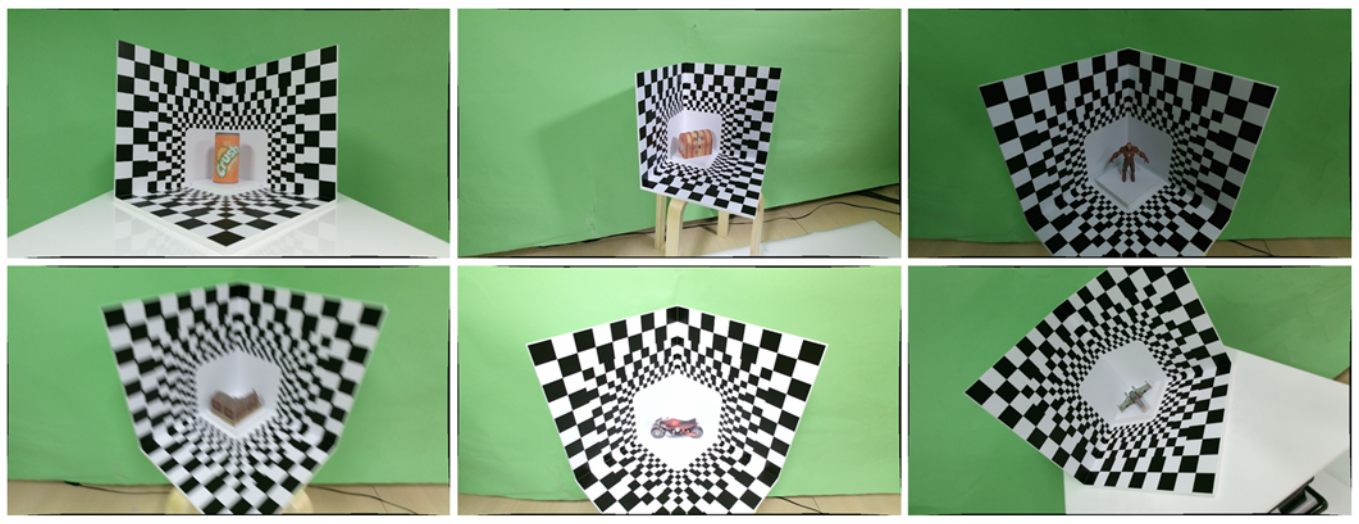

**Fig. 10.** Examples from OPT dataset [49].

TABLE II
EXPERIMENTAL RESULTS ON OPT DATASET. WE COMPARE WITH STATE-OF-THE-ART MONOCULAR 6DOF OBJECT TRACKING METHODS. THE BEST RESULTS ARE BOLD, THE SECOND-BEST RESULTS ARE UNDERLINED, AND THE ROTATIONALLY SYMMETRIC OBJECT IS COLORED RED.

| Method | Soda | Chest | Ironman | House | Bike | Jet | Average |
|---|---|---|---|---|---|---|---|
| IJCV12 [23] | 5.87 | 5.55 | 3.92 | 3.58 | 5.36 | 5.81 | 5.01 |
| TPAMI19 [25] | 8.86 | 11.76 | 11.99 | 10.15 | 11.90 | 13.22 | 11.31 |
| TIP20 [26] | 9.01 | 12.24 | 11.21 | 13.61 | 12.83 | 15.44 | 12.39 |
| JCST21 [36] | 9.00 | 14.92 | 13.44 | 13.60 | 12.85 | 10.64 | 12.41 |
| IJCV22 [28] | 15.64 | 16.30 | 17.41 | 16.36 | 13.02 | 15.64 | 15.73 |
| ECCV22 [30] | 4.20 | 9.20 | 3.26 | 4.05 | 7.63 | 8.65 | 6.16 |
| Ours (C) | 15.57 | 16.89 | 17.61 | 16.69 | 12.75 | 17.23 | 16.12 |
| Ours | **16.52** | **17.03** | **17.75** | **16.71** | **13.04** | **17.63** | **16.45** |

The performance of "Ours" shows improvements for all six objects, particularly for Soda object, which has rotational symmetry. The AUC score for Soda is 16.52, this result is consistent with that from the RBOT dataset, further validating the effectiveness of our method for tracking symmetric objects.

### C. Evaluation on BCOT Dataset

The BCOT dataset [50] is a real-world dataset featuring 20 objects and 404 sequences distributed across 22 scenes, consisting of a range of lighting conditions, camera configurations, movement behaviors, as well as indoor and outdoor settings. We assess tracking performance using ADD-$kd_m$ and $l$cm-$l$° metrics. For the ADD-$kd_m$ scores, we calculate the average vertex error $e_T$ for each frame using (29) and determine the percentage of frames where $e_T < kd_m$. In the case of $l$cm-$l$° scores, we apply (28) to evaluate translation error $e_t$ and rotation error $e_R$ for each frame, subsequently counting the frames where $e_t < l$cm and $e_R < l$°. If $e_t > 5$cm or $e_R > 5$°, we reset the object pose using the ground-truth pose label. Table III displays the performance of our method in comparison to advanced optimization-based monocular tracking methods.

Our method demonstrates superior performance on the $l$cm-$l$° and ADD-0.1$d_m$ metrics. A significant advantage of our approach is that it requires approximately 2,000 fewer pose reset times compared to state-of-the-art methods. Since color statistics are re-updated after pose resets, the use of pose resets may positively affect the tracking scores of subsequent frames on more precise metrics (e.g., ADD-0.02$d_m$ and ADD-0.05$d_m$). Consequently, we conducted tests and comparisons on the BCOT dataset without implementing any pose resets. As shown in Table IV, our method outperforms all competing methods across all metrics.

TABLE III
EXPERIMENTAL RESULTS ON BCOT DATASET. WE COMPARE WITH STATE-OF-THE-ART OPTIMIZATION-BASED MONOCULAR 6DOF OBJECT TRACKING METHODS. THE BEST RESULTS ARE BOLD, THE SECOND-BEST RESULTS ARE UNDERLINED.

| Method | ADD- | | | cm-degree | | Reset times |
|---|---|---|---|---|---|---|
| | $0.02d_m$ | $0.05d_m$ | $0.1d_m$ | 2-2 | 5-5 | |
| TPAMI19 [25] | 11.7 | 31.6 | 57.1 | 40.8 | 77.1 | - |
| TVCG22 [27] | **15.6** | 39.8 | 66.1 | 51.4 | 87.1 | - |
| IJCV22 [28] | 12.5 | 49.4 | 82.1 | 53.6 | 93.1 | 8548 |
| ECCV22 [30] | <u>15.3</u> | **52.1** | <u>82.7</u> | <u>63.2</u> | 93.8 | 7789 |
| Ours (C) | 14.0 | 51.1 | <u>82.7</u> | 62.5 | <u>94.0</u> | <u>7514</u> |
| Ours | 14.1 | <u>51.3</u> | **82.9** | **63.4** | **95.4** | **5872** |

TABLE IV
EXPERIMENTAL RESULTS ON BCOT DATASET WITHOUT POSE RESET.

| Method | ADD- | | | cm-degree | |
|---|---|---|---|---|---|
| | $0.02d_m$ | $0.05d_m$ | $0.1d_m$ | 2-2 | 5-5 |
| IJCV22 [28] | 9.2 | <u>36.2</u> | <u>60.7</u> | 41.9 | <u>69.6</u> |
| ECCV22 [30] | <u>10.5</u> | 34.5 | 54.2 | <u>43.7</u> | 61.4 |
| Ours | **10.6** | **38.0** | **61.5** | **48.4** | **71.0** |

TABLE V
AVERAGE RUNTIMES PER FRAME, HARDWARE USAGE AND AVERAGE TRACKING SUCCESS RATE ON RBOT DATASET FOR ADVANCED METHODS.

| Method | Hardware | | Runtime (ms) | Tracking success rate (%) |
|---|---|---|---|---|
| | CPU | GPU | | |
| TPAMI19 [25] | √ | √ | 15.5 ~ 21.8 | 72.8 |
| TIP20 [26] | √ | √ | 41.2 | 76.5 |
| TIE21 [39] | √ | √ | 6.9 | 77.7 |
| JCST21 [36] | √ | √ | 32.1 | 81.1 |
| TCSVT21 [35] | √ | √ | 40 ~ 50 | 85.6 |
| TVCG22 [27] | √ | √ | 33.1 | 84.8 |
| IJCV22 [28] | √ | × | **1.1** | 90.9 |
| ECCV22 [30] | √ | × | 13.6 | 92.2 |
| TIE23 [29] | √ | √ | 4 | 91.6 |
| ICCV23 [42] | √ | √ | 133.7 | <u>93.3</u> |
| AEI24 [37] | √ | × | 14.8 | 92.5 |
| Ours (C) | √ | × | <u>2.7</u> | 91.9 |
| Ours | √ | × | 6.6 | **95.4** |

### *D. Runtime Analysis*

Table V lists the runtime, hardware usage and the average tracking success rates of our method and compared methods on the RBOT dataset. For the latest advanced open-source methods [28], [30], and [42], we ran them on the same laptop using the recommended parameters. For the other methods, we referred to the hardware specifications and testing speeds reported in their respective publications. On our device, the average running time of "Ours (C)" is 2.7ms, which is only slower around 1.6ms than [28], yet achieves nearly 1% improvement of average tracking success rate. The running time of "Ours" is 6.6ms, which is slower than [28], but it provides a higher tracking success rate by 4.5%. When compared with [30], which achieves 13.6ms, "Ours" is faster and more accurate. For recent advanced optimization-based methods [37], "Ours" achieves faster speed and higher tracking success rate. Compared to the deep learning-based method [42], our method is approximately 20× faster. The comparisons prove that our method is efficient and robust enough to satisfy the requirements of real-world AR applications. Additionally, we have also evaluated the storage requirements of our method. Our method uses approximately 200MB of memory during runtime, and each object requires 21.5MB of external storage for the sparse viewpoint model.

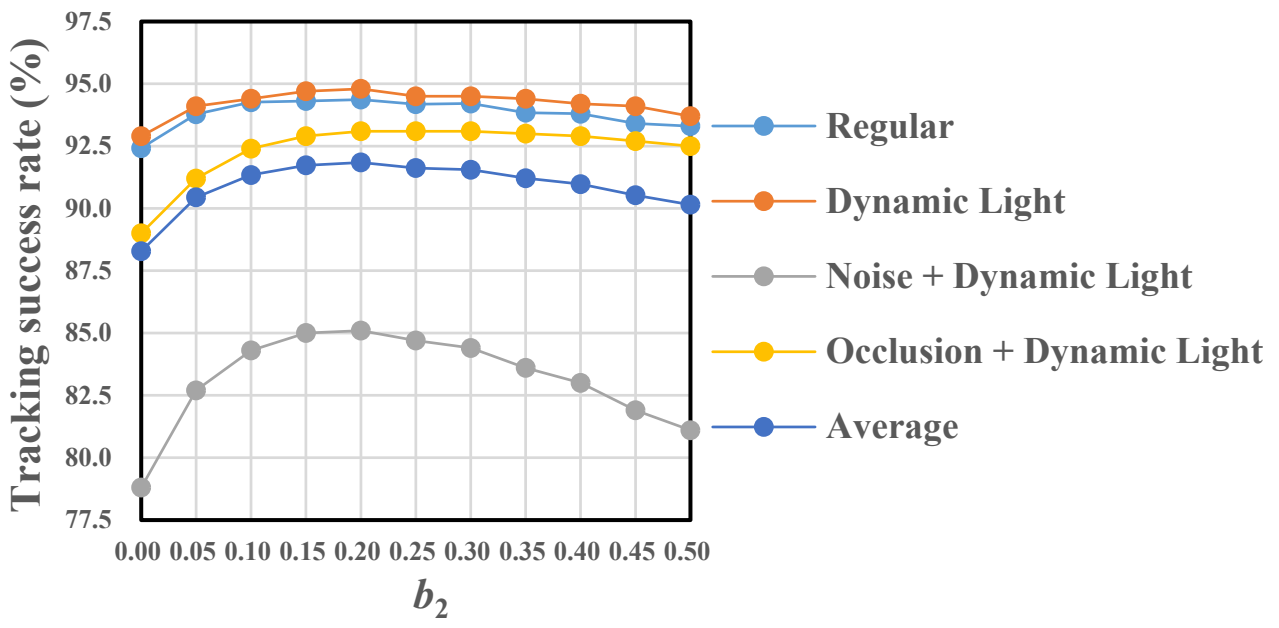

**Fig. 11.** Tracking success rates for different values of $b_2$ on RBOT dataset.

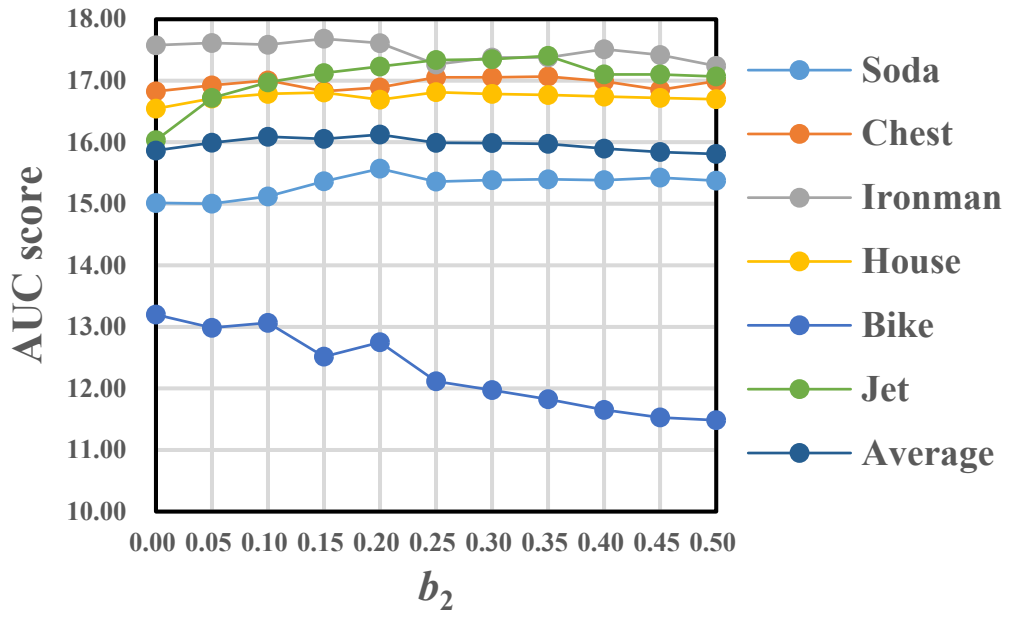

**Fig. 12.** AUC scores for different values of $b_2$ on OPT dataset.

### *E. Ablation Studies*

In the following ablation studies, we conduct experiments on the RBOT and OPT datasets. We analyze several parameters, including those related to the mixed distribution, the sampling interval angle and searching range of the fan-shaped area, the weight of the contour energy function, and the patch-wise interior correspondence uncertainty, to assess their influence on tracking performance.

*1) The Parameter of The Mixed Distribution:* As described in section III, parameter $b_2$ determines the proportion of Gaussian distribution and uniform distribution in the mixed probability model. For instance, when $b_2 = 0$, the mixed distribution would degenerate to a Gaussian distribution. Fig. 11 and Fig. 12 illustrate that $b_2$ has a more significant impact on tracking success rates compared to AUC scores. Interestingly, the highest AUC and tracking success rates are achieved when $b_2$ is set to 0.20 for both datasets.

Images in RBOT dataset exhibit heavily cluttered background and noisy points. As a result, the means of the distribution are susceptible to noise. Appropriately increasing $b_2$ raises the proportion of the uniform distribution, which helps mitigate the effects of outliers. In contrast, for OPT dataset, the background is nearly pure white, which makes the outlier suppression strategy less effective in improving performance. Additionally, as illustrated in Fig. 11 and Fig. 12, high value of $b_2$ makes pose optimization difficult to converge, ultimately leading to a decrease in tracking success rates.

*2) Fan-shaped Contour Points Search Strategy:* As described in section IV, we employ a coarse-to-fine fan-shaped search strategy to build contour correspondences. The sampling interval angle $\Delta A_{int}$ and the angle of fan-shaped region $A_{reg}$ significantly influence the tracking success rates.

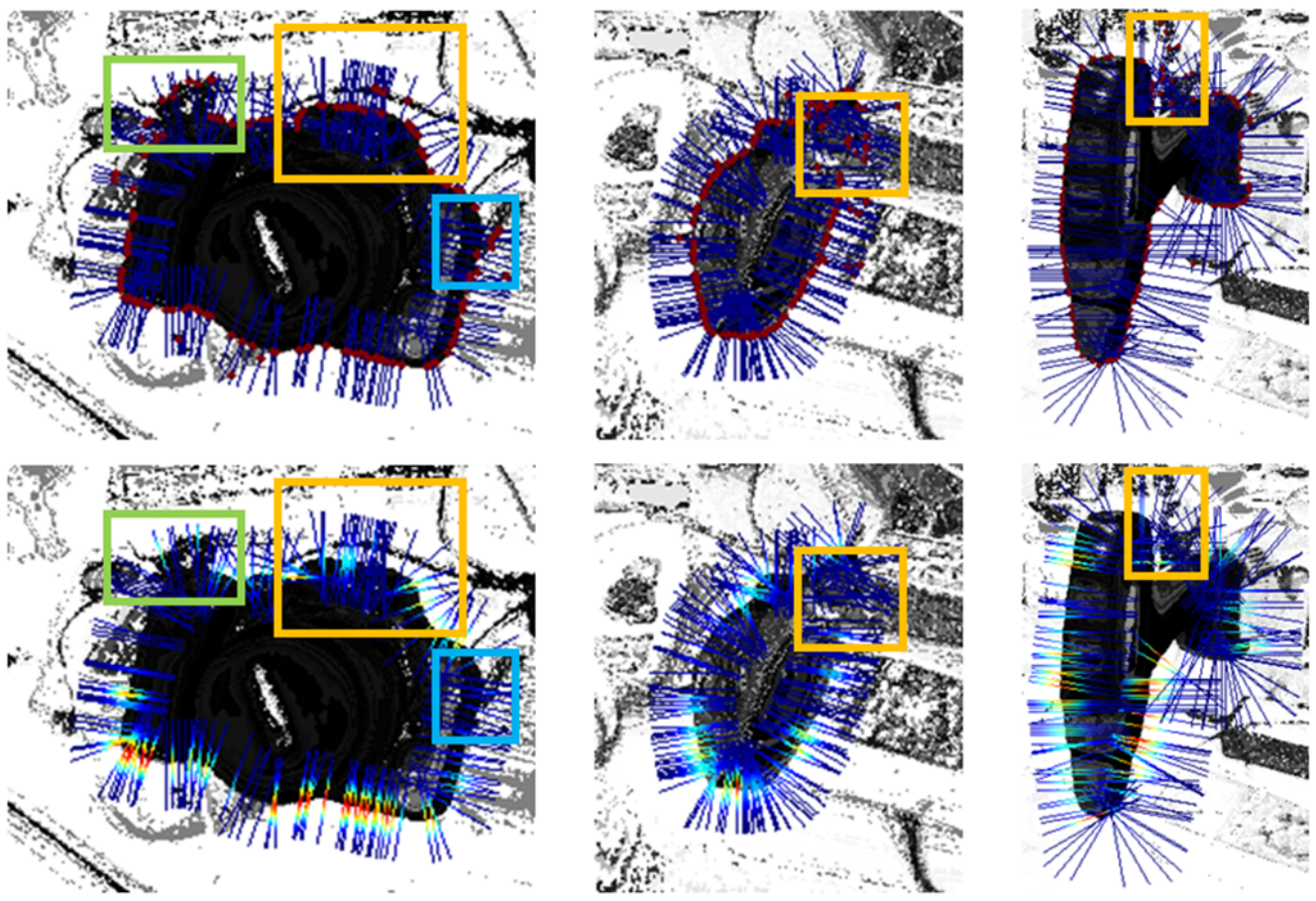

(a) Camera (b) Glue (c) Driller

**Fig. 13.** Qualitative analysis of different data association strategies. For each object, the first row displays the results of previous contour point search strategy, while the second row visualizes our contour weights, with the box area indicating the error scenarios from previous strategy.

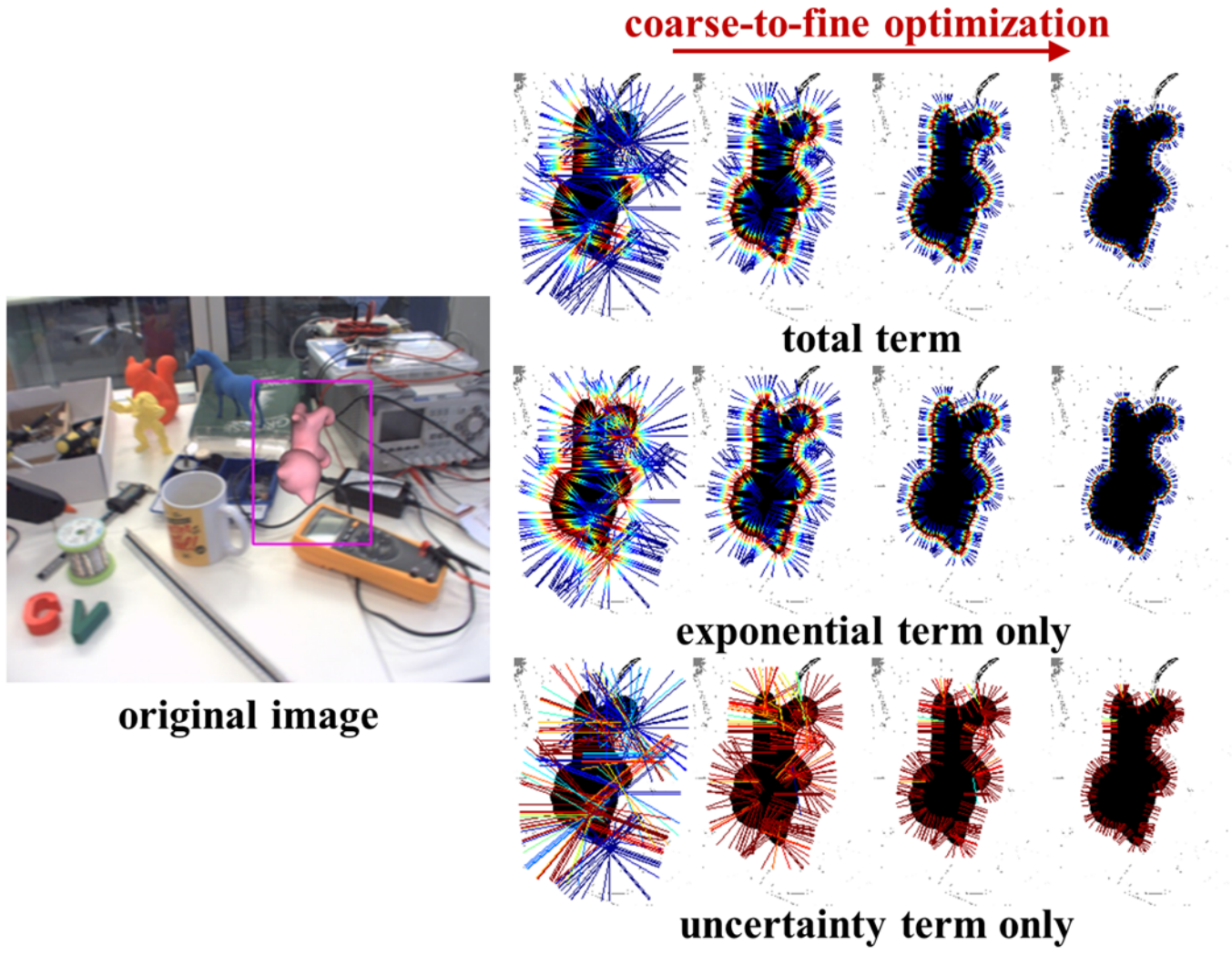

**Fig. 14.** Qualitative analysis of our contour weight.

TABLE VI
TRACKING SUCCESS RATES (IN %) FOR DIFFERENT COMBINATIONS OF SAMPLING INTERVALS AND SEARCH REGION ANGLES ON RBOT DATASET.

| $\Delta A_{int}$ | 20° | | | 10° | | | 5° | | |
|---|---|---|---|---|---|---|---|---|---|
| $\{A_{reg}\}$ | {0°, 0°, 0°, 0°} | {120°, 80°, 40°, 0°} | {240°, 160°, 80°, 0°} | {0°, 0°, 0°, 0°} | {60°, 40°, 20°, 0°} | {120°, 80°, 40°, 0°} | {0°, 0°, 0°, 0°} | {30°, 20°, 10°, 0°} | {60°, 40°, 20°, 0°} |
| Reg. | 93.6 | 93.9 | 91.6 | 93.9 | **94.4** | 94.1 | 93.9 | 94.2 | **94.4** |
| Dyn. | 93.8 | 94.1 | 92.1 | 94.3 | **94.8** | 94.6 | 94.3 | 94.6 | 94.7 |
| Noi. | 84.0 | 83.2 | 76.0 | 84.1 | **85.1** | 83.1 | 84.3 | **85.1** | 84.8 |
| Occ. | 92.1 | 92.8 | 90.2 | 92.3 | 93.1 | 93.1 | 92.3 | 92.7 | **93.2** |
| Avg. | 90.9 | 91.0 | 87.5 | 91.2 | **91.9** | 91.2 | 91.2 | 91.7 | 91.8 |

In our experiments, the coarse-to-fine strategy has four stages to converge. Three $\Delta A_{int}$ of 20°, 10° and 5° are tested. For each $\Delta A_{int}$, we evaluate three sets of $\{A_{reg}\}$. {0°, 0°, 0°, 0°} corresponds to searching the contour points solely along the normal direction, $\{6\Delta A_{int}, 4\Delta A_{int}, 2\Delta A_{int}, 0°\}$ means a moderate fan-shaped search region, and $\{12\Delta A_{int}, 8\Delta A_{int}, 4\Delta A_{int}, 0°\}$ indicates a larger search region. As shown in Table VI, the highest average tracking success rate is achieved when $\Delta A_{int}$ is set to 10° and $\{A_{reg}\}$ is set to {60°, 40°, 20°, 0°}. When $\{A_{reg}\}$ is set to {0°, 0°, 0°, 0°}, our method degenerates into RAPID-like methods where correspondences are searched only along normal search lines. However, thanks to the noise uncertainty and mixed probability model, our contour-based method can still achieve relatively high tracking success rates. We also observe that excessively large search ranges or dense sampling may not necessarily enhance performance.

We also visualize the situation when $\{A_{reg}\}$ is set to {0°, 0°, 0°, 0°}, as shown in the Fig. 13. We use red filled circles to indicate the contour points only searched by the normal lines. For regions where point-to-point correspondences are prone to incorrect, such as regions with high object contour curvature and image interference, our point-to-distribution strategy eliminates their impact by reducing their weights.

*3) The Weight of Contour Energy:* In our contour-based method, we utilize a mixed probability to model the distribution of image contour correspondences along the normal search line.

TABLE VII
TRACKING SUCCESS RATES (IN %) FOR DIFFERENT COMBINATIONS OF CONTOUR WEIGHT PARTS ON RBOT DATASET.

| Uncertainty term | | Exponential term | Tracking scene | | | | |
|---|---|---|---|---|---|---|---|
| $1/\sigma_{shp,i}^2$ | $1/\sigma_{noi,i}^2$ | $e^{-\beta r_{cnt,i}^2}$ | Reg. | Dyn. | Noi. | Occ. | Avg. |
| × | × | × | 87.0 | 87.9 | 67.2 | 81.1 | 80.8 |
| √ | × | × | 89.3 | 90.1 | 71.3 | 84.4 | 83.8 |
| × | √ | × | 90.6 | 91.4 | 73.1 | 86.4 | 85.4 |
| × | × | √ | 92.1 | 92.5 | 77.3 | 89.1 | 87.8 |
| √ | √ | × | 92.4 | 92.9 | 78.8 | 89.0 | 88.3 |
| √ | × | √ | 93.3 | 93.7 | 81.5 | 91.2 | 89.9 |
| × | √ | √ | 93.6 | 94.0 | 81.4 | 91.9 | 90.2 |
| √ | √ | √ | **94.4** | **94.8** | **85.1** | **93.1** | **91.9** |

TABLE VIII
TRACKING SUCCESS RATES (IN %) WITH AND WITHOUT PATCH-WISE INTERIOR CORRESPONDENCE UNCERTAINTY ON RBOT DATASET.

| $c_{in,i}$ | Reg. | Dyn. | Noi. | Occ. | Avg. |
|---|---|---|---|---|---|
| × | 96.7 | 96.6 | 91.5 | 95.3 | 95.0 |
| √ | **97.1** | **97.0** | **92.0** | **95.6** | **95.4** |

Based on this model, we derive a robust weighted energy function $E_{cnt}$ which consists of the uncertainty term $1/\sigma_i^2$ (a product of shape uncertainty $1/\sigma_{shp,i}^2$ and noise uncertainty $1/\sigma_{noi,i}^2$ ) and an exponential term $e^{-\beta r_{cnt,i}^2}$. Each of them takes significant effects. The ablation results on RBOT dataset are shown in Table VII. It can be observed that the energy function $E_{cnt}$ without weights degenerates into a general least square problem, resulting in the lowest tracking success rates. Individually considering these terms can significantly improve the average tracking success rates by 3%, 4.6%, and 7%. While, combining all terms together will further improve performance, and achieves the highest average tracking success rate 91.9%.

Fig. 14 illustrates the important role played by those contour weights. For the uncertainty term $1/\sigma_i^2$, it suppresses correspondences which are located in error-prone areas, such as high contour curvature. And for the exponential term $e^{-\beta r_{cnt,i}^{2}}$, it suppresses outliers that are far from the mean of distribution. Taken together, these two terms ensure the tracking robustness.

*4) The Patch-wise Interior Correspondence Uncertainty:* Table VIII demonstrates the effectiveness of the patch-wise interior correspondence uncertainty $c_{in,i}$ we introduced. Using $c_{in,i}$ achieves higher tracking success rates for all four cases than not using it, with an average improvement of 0.4%.

### *F. Applications of Augmented Reality Assembly Guidance*

*1) Evaluation on Real Scenarios:* Besides evaluating on public datasets, we perform additional experiments on AR assembly guidance scenario to validate the effectiveness of our approach. In the experiments, we used a self-developed AR headset equipped with a camera module and optical see-through displays. The computing unit is separate from the optical display module to optimize thermal management and is connected to the AR headset via USB 3.0. The camera SDK and our object tracking algorithm are compiled into dynamic link libraries (DLLs). The integration of the algorithm and the workflow are illustrated in Fig. 15.

In this application, our method tracks the pose of texture-less industrial parts in real time, such as the Electric machine and Battery box considered in this paper. Fig. 16(a) and Fig. 16(b) show the AR assembly guidance system and the coordinates involved in tracking. Based on the obtained real-time pose of the assembly, it can estimate the relative 6DoF pose between the industrial parts hold on the operator (see the blue models in Fig. 16(c) and Fig. 16(e)) and the target assembly position rendered by computer (see the red models in Fig. 16(c) and Fig. 16(e)), and then judge whether the assembly result is correct or whether the current assembly step is completed. The relative 6DoF pose $\mathbf{T}_{MT}$ of the target position with respect to the manipulated object can be computed by

$$\mathbf{T}_{MT} = \left(\mathbf{T}_{WC}\mathbf{T}_{CM}\right)^{-1}\mathbf{T}_{WT}\,. \tag{30}$$

Here, the pose $\mathbf{T}_{WC}$ between the camera to world coordinate system can be obtained by the SLAM module on the HMD, transformation matrix $\mathbf{T}_{CM}$ of the manipulated object relative to the camera is calculated in real-time using our algorithm. While $\mathbf{T}_{WT}$, the target assembly position in the world coordinate system, is known and can be rendered by the 3D rendering engine Unity3D. When the translation error of $\mathbf{T}_{MT}$ is less than 1cm and the rotation error of $\mathbf{T}_{MT}$ is within 1°, both the guiding arrow and the target model change from red to green, indicating to the operator that the assembly is correct, as shown in Fig. 16(d) and Fig. 16(f). The 3D models we used are depicted in Fig. 17.

To further evaluate the tracking robustness of our algorithm, we create challenging scenes featuring cluttered backgrounds, object with rotational symmetry, occlusion, and large displacement. Fig. 18(a) demonstrates the robust tracking results in a cluttered background. Robust contour weight helps to reduce the influence of wrong contour correspondences. This result is also validated on public dataset, as shown in Table I.

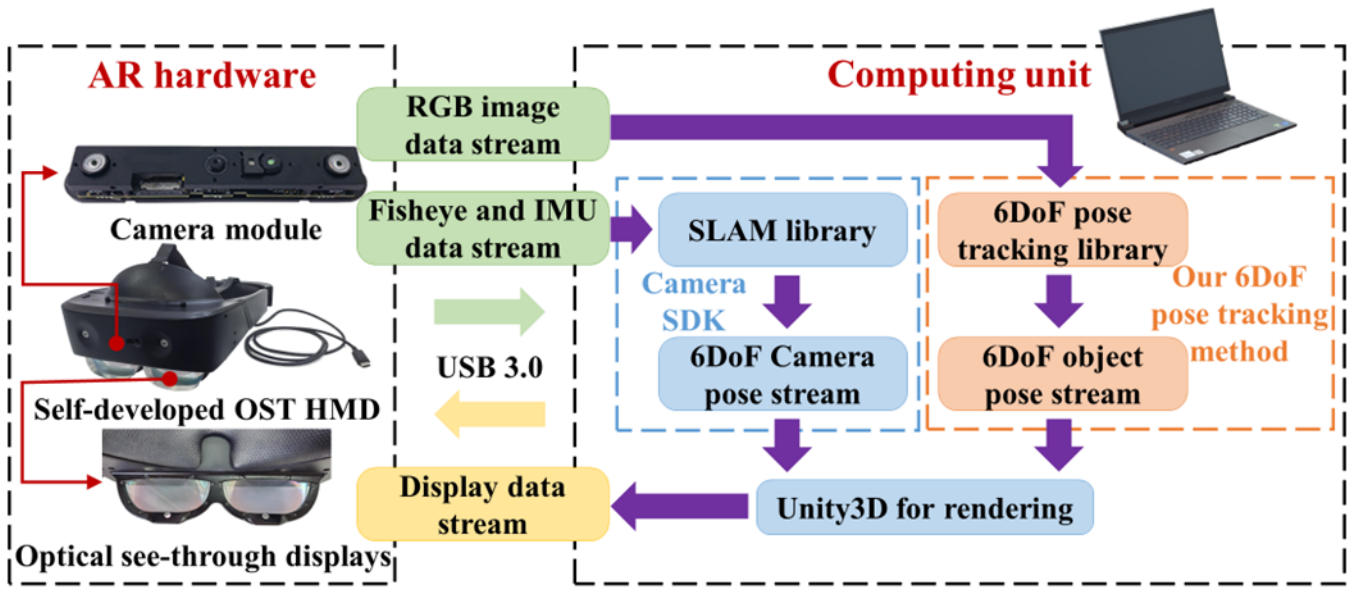

**Fig. 15.** The integration and workflow of our pose tracking method with self-developed AR hardware.

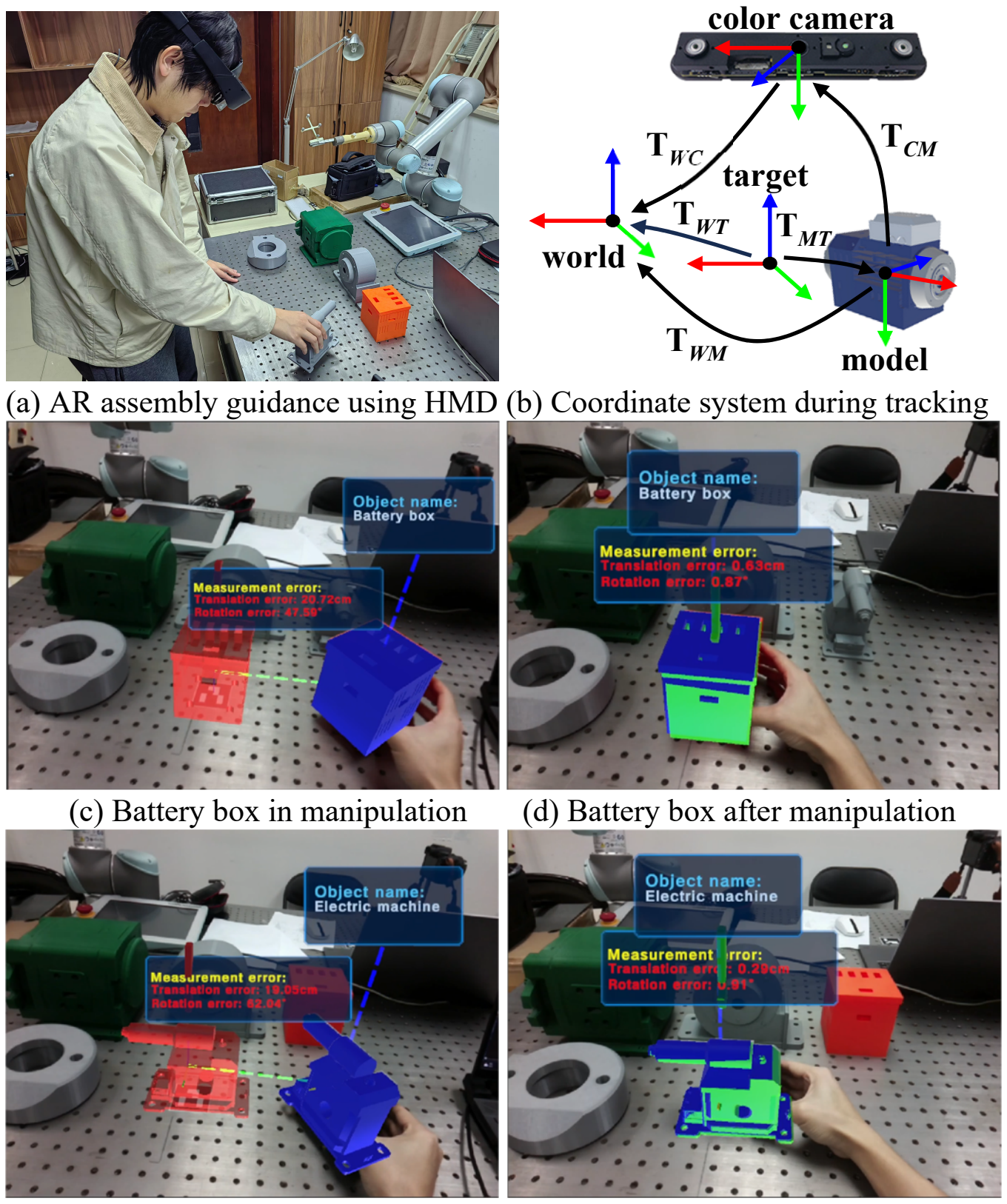

(a) AR assembly guidance using HMD (b) Coordinate system during tracking

(c) Battery box in manipulation (d) Battery box after manipulation

(e) Electric machine in manipulation (f) Electric machine after manipulation

**Fig. 16.** Application of AR assembly guidance.

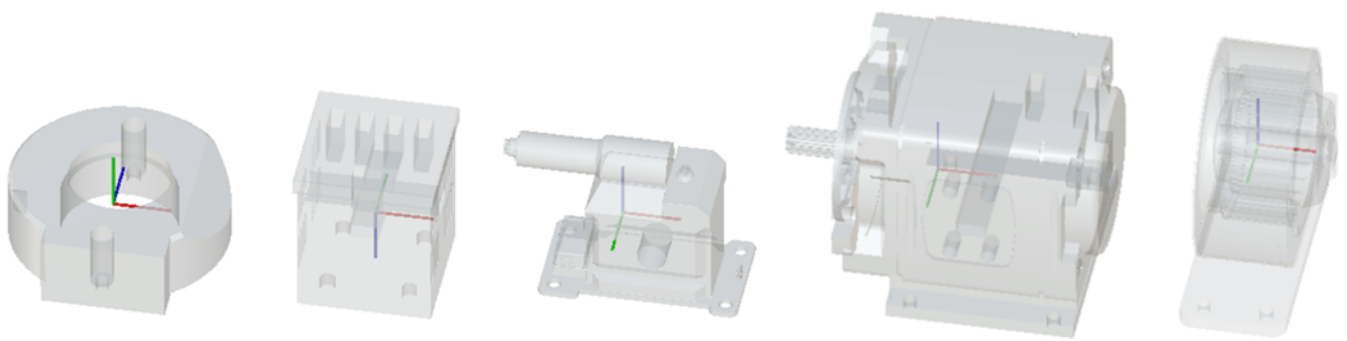

(a) Base (b) Battery box (c) Electric machine (d) Engine (e) Industrial fan

**Fig. 17.** The 3D models for our application.

In Fig. 18(b), we visualize the tracking performance for objects with rotational symmetry, such as the Battery box and Industrial fan which exhibits rotationally symmetric appearance after part of its region is occluded. It is worth mentioning that handling ambiguous contours is crucial, as many industrial parts naturally exhibit symmetric characteristics. As shown in Fig. 19, the comparison between “Ours (C)” and “Ours” indicates that using interior correspondences helps address the ambiguous contour problem and enables more robust tracking.

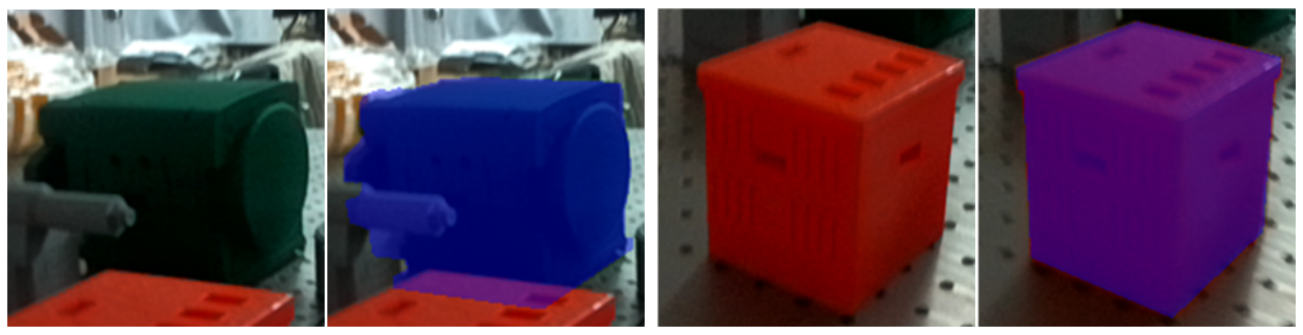

(a) Cluttered background (b) Object with rotational symmetry

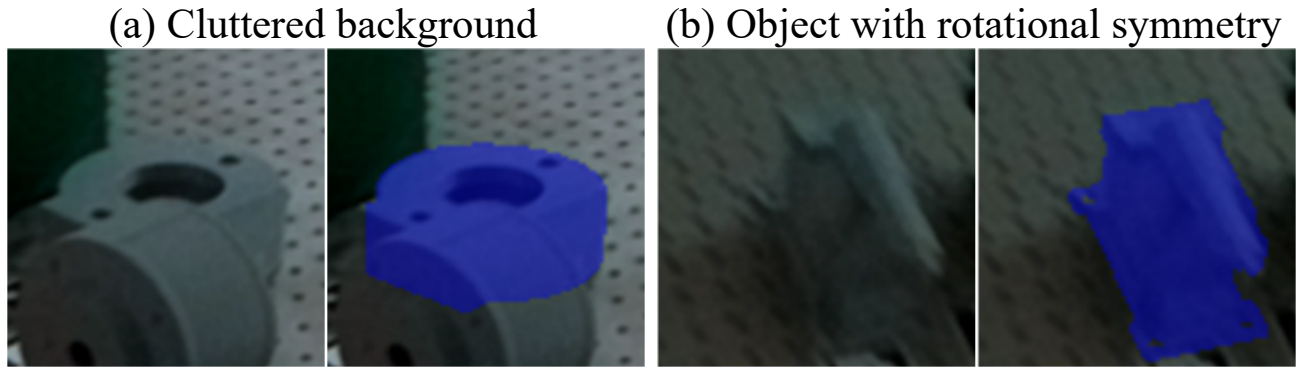

(c) Partial occlusion (d) Large displacement

**Fig. 18.** Performances of our method in challenging situations. For each scene, the first image is the original image, the second image is our tracking result, we use estimated pose and blue color to render AR image.

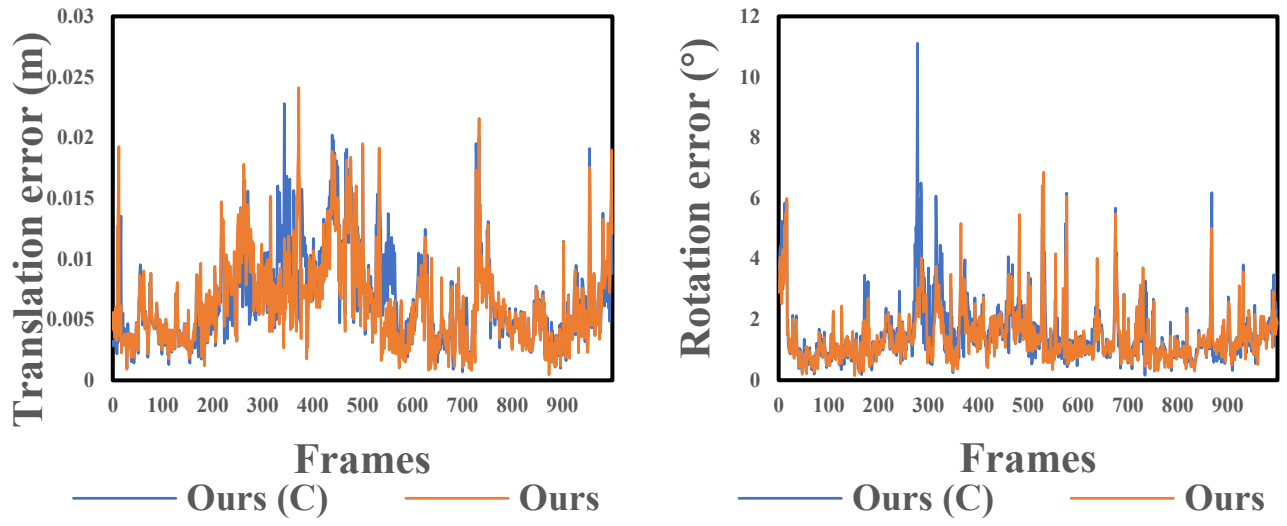

(a) Error curves for Battery box

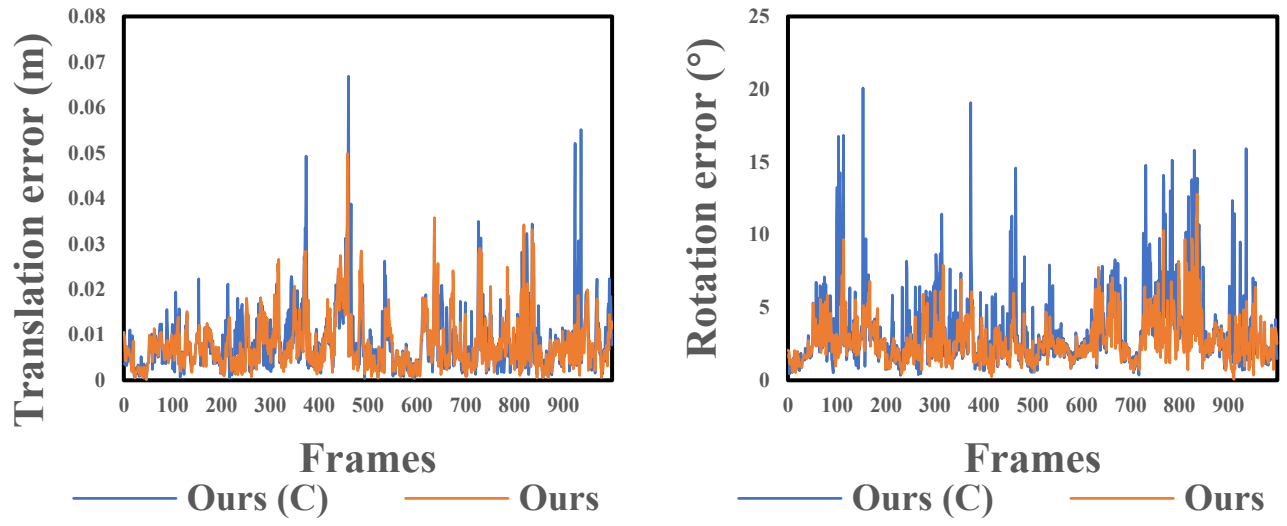

(b) Error curves for Industrial fan

**Fig. 19.** Error curves of two manipulated objects with rotational symmetry.

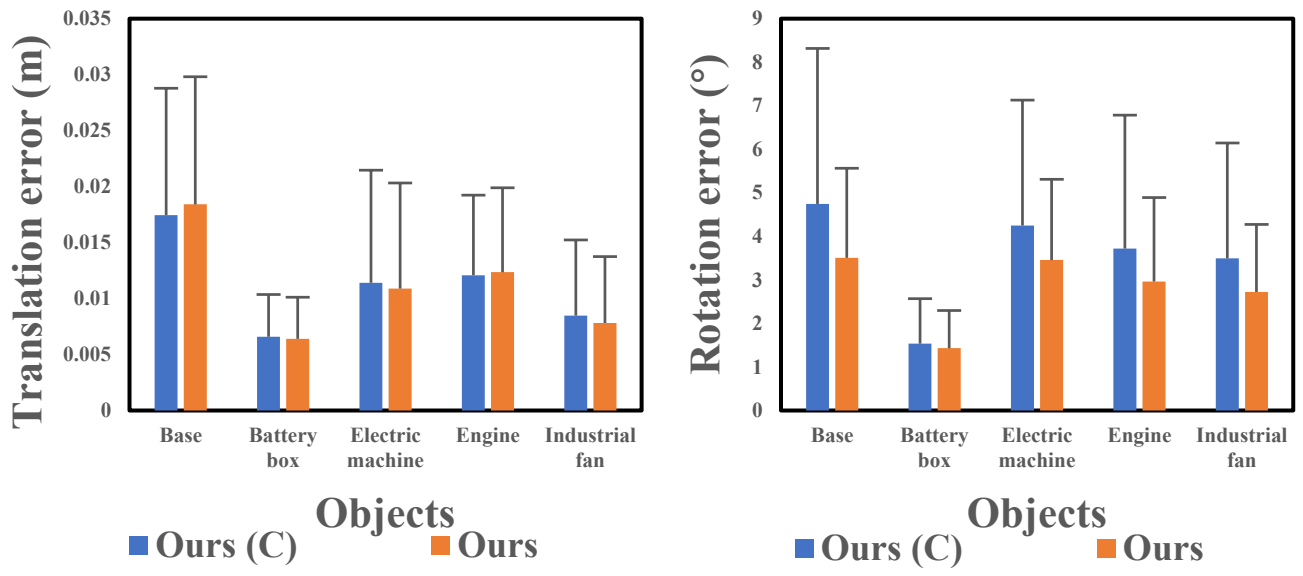

**Fig. 20.** The mean and standard deviation of translation and rotation errors of our methods.

During the assembly process, partial occlusion is unavoidable. Fig. 18(c) demonstrates the tracking results under this condition. Our robust weights help suppress erroneous correspondences in occluded parts, thus enhancing the tracking performance.

In Fig. 18(d), due to the rapid movement of the operator's head or hand, the tracked object has large displacement. However, thanks to efficient interior correspondences, our tracker is able to maintain successful tracking.

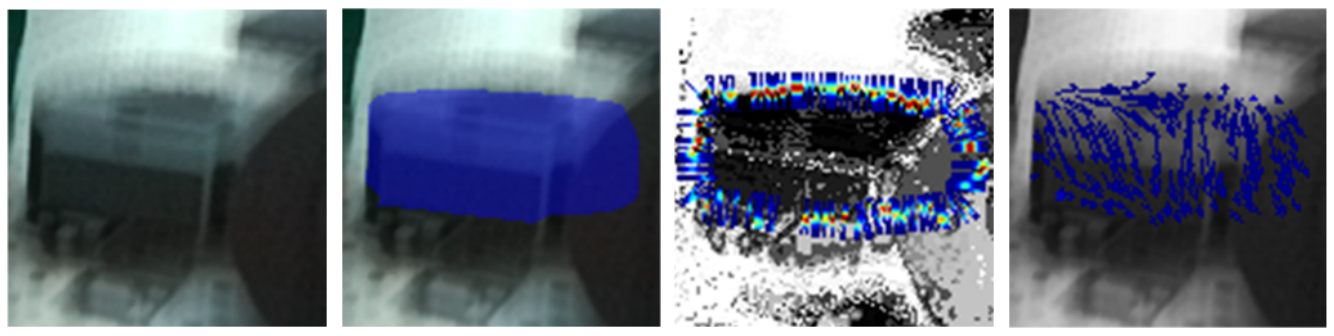

(a) Severe motion blur

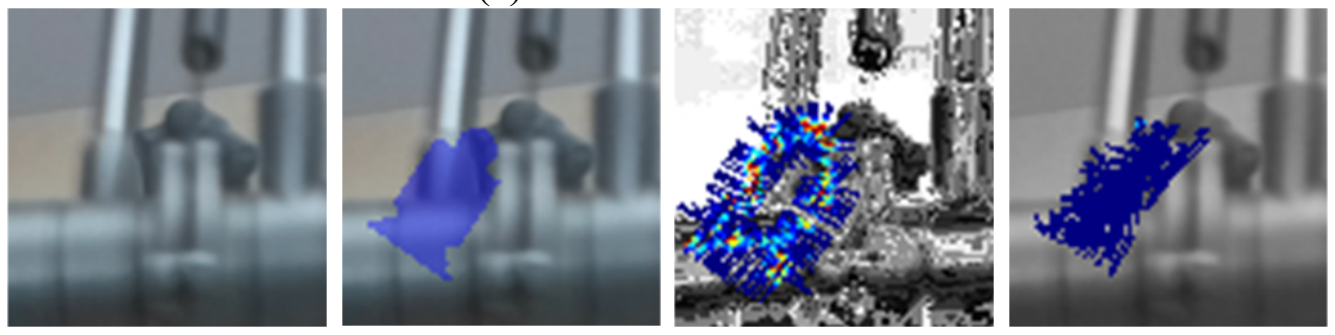

(b) Severe occlusion

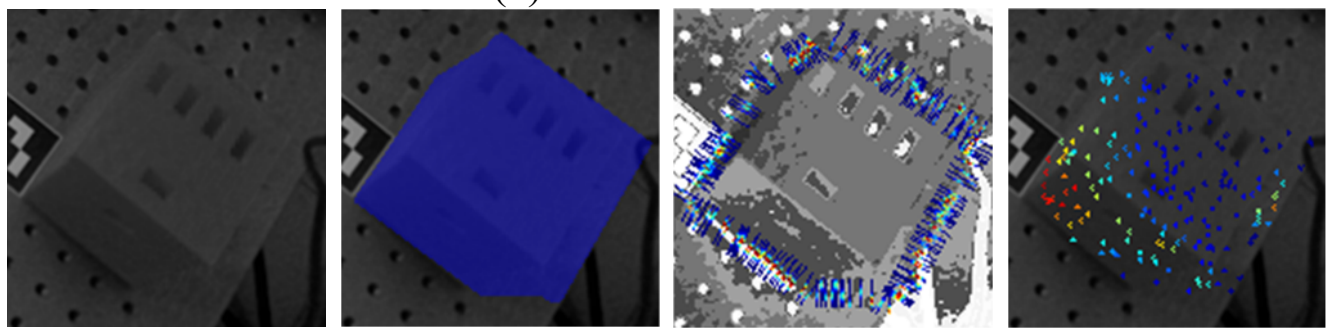

(c) Extremely low-contrast scene

**Fig. 21.** Failure cases. For each case, the tracking result, contour correspondence, and interior correspondence are visualized.

We measure the mean and standard deviation of translation and rotation errors of five manipulated objects in the scene, as shown in Fig. 20, utilizing interior information of the object significantly improves the accuracy of rotation estimation.

*2) Discussion of Practical Application Scenarios:* In static scenarios—where the object to be assembled is fixed within the scene—such as automotive manufacturing and household appliance installation, the SLAM algorithm is employed to determine the real-time pose of the camera within the scene. Based on this, our tracking method can acquire the 6DoF pose of the operational components relative to the scene.

For non-static scenarios—where the components to be assembled are freely movable—such as electronic device or furniture assembly, our algorithm can simultaneously process the pose of multiple components. Specifically, the algorithm can track both assembled components and components to be assembled separately, and compute their relative poses.

The AR guidance includes dynamic instructions, warnings, and visual cues via 2D/3D rendering. Experiments confirm our algorithm's high accuracy, real-time performance, and robustness to variations in lighting, occlusion, and scene changes, ensuring reliable AR assembly guidance.

## G. Limitations

While our 6DoF object tracking methods demonstrate high robustness, there is still room to improve in the future.

*1) Severe Motion Blur:* Caused by slow shutter speed, fast object motion, or camera shake, motion blur results in ghosted contours and compromised internal features, leading to inaccurate correspondences and pose drift (Fig. 21(a)). Improvements include image deblurring, multi-sensor fusion (e.g., using multi-view color cameras to introduce extra constraints), or using event cameras to avoid motion blur.

*2) Severe Occlusion:* Severe occlusion causes incorrect associations between the object and occluding elements, leading to tracking failure (Fig. 21(b)). An occlusion-aware strategy, such as instance segmentation using deep learning, could detect non-occluded regions for correspondence building. Alternatively, depth cameras could differentiate occlusions based on depth differences.

*3) Extremely Low Color Contrast:* When the object and background share similar colors, segmentation fails, violating the contour-based tracking assumption (Fig. 21(c)). Incorporating image gradient information or training semantic segmentation networks could improve contour robustness.

## VI. Conclusion

In this paper, we propose a robust monocular 6DoF object pose tracking method considering both contour and interior information for efficient AR assembly guidance. By modeling the shape and noise uncertainty of local contours as mixed probability distributions, our contour-based method performs exceptionally well, particularly in tracking under noisy conditions. By introducing efficient interior correspondences, our multi-feature-based tracker significantly outperforms existing monocular tracking methods, especially when dealing with rotationally symmetric objects, achieving 100 FPS using only the CPU.